\documentclass[runningheads]{llncs}
\usepackage{graphicx}

\usepackage{amsmath}
\usepackage{subcaption}
\begin{document}
%
%\title{Photography support system considering \\angle of view by eliminating false detection \\of face detection and undetected interpolation using an omni-directional camera and \\facial expression recognition}
\title{Selfie Taking with Facial Expression Recognition Using Omni-directional Camera}
\author{Kazutaka Kiuchi\and
Shimpei Imamura\and
Norihiko Kawai\orcidID{0000-0002-7859-8407}}
\authorrunning{K. Kiuchi et al.}
\institute{Graduate School of Information Science and Technology,\\ Osaka Institute of Technology 
\\1-79-1 Kitayama, Hirakata, Osaka 573-0196, Japan\\
\email{norihiko.kawai@oit.ac.jp}}
\maketitle              % typeset the header of the contribution
\begin{abstract}
Recent studies have shown that visually impaired people have desires to take selfies in the same way as sighted people do to record their photos and share them with others. Although support applications using sound and vibration have been developed to help visually impaired people take selfies using smartphone cameras, it is still difficult to capture everyone in the angle of view, and it is also difficult to confirm that they all have good expressions in the photo. To mitigate these issues, we propose a method to take selfies with multiple people using an omni-directional camera. Specifically, a user takes a few seconds of video with an omni-directional camera, followed by face detection on all frames. The proposed method then eliminates false face detections and complements undetected ones considering the consistency across all frames. After performing facial expression recognition on all the frames, the proposed method finally extracts the frame in which the participants are happiest, and generates a perspective projection image in which all the participants are in the angle of view from the omni-directional frame. In experiments, we use several scenes with different number of people taken to demonstrate the effectiveness of the proposed method.

\end{abstract}
\section{Introduction}
According to a survey reported in 2011\cite{jayant2011supporting}, about 71\% of visually impaired people had recently taken photos. This suggests that many visually impaired people have the same desire to take photos and share them with others in the same way as sighted people do. On the other hand, a survey published in 2021 reported that about 90\% of visually impaired people use iPhone's VoiceOver app, which has a function that tells how many people are within the field of view of the camera when taking a photo of people, allowing the visually impaired to take photos of people. However, since this function recognizes a face as long as 70\% of the face is in the frame, when a person takes a selfie with multiple people, they may not be in the frame. In addition, it is difficult for them to distinguish whether or not the photographed persons have good facial expressions.

In this study, focusing on a situation in which visually impaired people take selfies, we propose a method that enables to take selfies with multiple people using an omni-directional camera. Specifically, a user first takes a few seconds of video using an omni-directional camera. Next, the proposed method performs face detection on all frames, and then eliminates false face detections and complements undetected ones considering the consistency across all frames. After performing facial expression recognition on all the frames, the proposed method finally extracts the frame in which the participants are happiest, and generates a perspective projection image in which all the participants are in the angle of view from the omni-directional frame.

\section{Related Work}
This section introduces photography systems that support the visually impaired and methods using omni-directional images related to this study. Several methods have been proposed to assist visually impaired people in taking photographs. Jayant et al. proposed EasySnap\cite{jayant2011supporting,white2010easysnap}, which enables users to adjust the composition, zoom level, and brightness of the subject according to voice instructions. They also proposed PortraitFramer\cite{jayant2011supporting}, which enables users to be informed of the positions of faces by vibration when the smartphone screen is touched, allowing the users to move the camera to achieve the desired composition. V{\'a}zquez et al.\cite{vazquez2014assisted} proposed to automatically adjust the camera roll, and Balata et al.\cite{balata2015blindcamera} provide feedback to users on compositions of their photos using center and golden ratio compositions. These methods enable users to confirm the composition before taking photos. 
In addition, services and systems have been proposed to support visually impaired people \cite{bigham2010vizwiz,BeMyEyes,EnvisionAI,TapTapSee,OrCamMyEye2,vOICe,kutiyanawala2011teleassistance,kacorri2017people,cutter2015towards} in their daily lives. Although these help visually impaired people obtain information through photographs, none of them are intended to take selfies.

%%%木内案
%これらのカメラやスマートフォンなどのウェアラブル端末を利用する方法は，あらかじめ撮影対象の位置を把握し，カメラの画角に入るように撮影する必要がある．このようなシステムを扱う為には，ユーザーはある程度の訓練を受ける必要がある．また，Ricardoらの調査によると，視覚障害者にとって「スマートフォンを右に45°回せ」などの音声指示に確実に従うことは難しく，振動によるフィードバックが混乱の原因になると不快に感じる利用者もいることが述べられている．
%これらの問題を解決するために，カメラを撮影対象に向けることなく，撮影対象を画像に収めることが可能なデバイスである全方位カメラを用いる方法が提案されている．Jokelaらの全方位カメラの使用感の調査結果から，自撮りや集合写真を撮ることは全方位カメラの重要な用途であると言える．しかし，多くの情報が含まれる全方位画像は時に面白みのない写真になることもあるとも述べており，全方位画像をVRで視聴することも優れた体験ではあるが現実的な問題が多いことも述べている．この問題に対して，岩村らが提案したVizPhotoは記録された音声情報と撮影日時に加え、物体検出により検出された物体のリストを用いて、Webインタフェース上で全方位画像を選択し、検出された物体から任意の物体を選択し、透視投影画像を出力している。この方法は、画角内の対象物を撮影することには成功しているが、自撮りを目的としたものではなく、人間の表情を考慮したものではない。また，全方位カメラを用いた自撮りの研究としてobataらは離れ離れになった家族とコミュニケーションを取るために食事中に全方位ビデオメッセージを送り合うシステムを提案している．本システムの機能の一つとして顔追跡機能による全方位動画中の注目すべき人物の顔を拡大した映像を全方位動画に合成している．しかし，この研究はビデオメッセージによるコミュニケーションに着目しており，表情の良い写真を撮影することを目的としていない．
As mentioned above, although many support systems have been proposed that use the common camera in smartphones, it is still difficult for the visually impaired to grasp the positions of objects to be photographed and adjust the camera so that they are included within the angle of view of the camera. To handle these systems, the users need to be trained to a certain degree. Gonzalez Penuela et al. \cite{gonzalez2022understanding} also reported that it is difficult for visually impaired people to reliably follow voice instructions such as "turn your smartphone 45° to the right", and that some users find the vibratory feedback confusing and uncomfortable. 

To solve these problems, methods using an omni-directional camera that can capture an image of an object without pointing the camera at the object has been proposed. Based on the results of Jokela et al.'s survey \cite{jokela2019people} on the usability of omni-directional camera, taking selfies and group photos is important use of an omni-directional camera. However, they reported that omni-directional images containing a lot of information can sometimes be uninteresting. To address this issue, Iwamura et al. \cite{iwamura2020visphoto} proposed VizPhoto, which selects an arbitrary object from the detected objects, and outputs a perspective projection image. Although this method succeeds in capturing the object within the angle of view, it does not aim to take selfies and does not take human facial expressions into consideration. As a study of selfies using an omni-directional camera, obata et al. \cite{obata2020asynchronous} proposed a system for sending omni-directional video messages to family members who are separated from each other during a meal in order to communicate with them. As one of the functions of this system, the face image of the person of interest in the omni-directional video is combined with the frontal view of the omni-directional video by a face tracking function to share both the face and the view the user are seeing. However, this research focuses on communication through video messages, and does not aim to take selfies with good facial expressions.
%To solve the problem, Iwamura et al. \cite{iwamura2020visphoto} proposed VizPhoto, which uses an omni-directional camera, allowing an object to be captured in the image without having to point the camera at the object. VizPhoto selects an omni-directional image on a web interface using a list of objects detected by object detection in addition to the recorded audio information and the shooting date and time, selects an arbitrary object from the detected objects, and outputs a perspective projection image. Although this method succeeds in capturing the object within the angle of view, it does not aim to take selfies and does not take human facial expressions into consideration. 
%%%先生の英語の和訳
%この問題を解決するために、岩村らは、全方位カメラを用いることで、カメラを被写体に向けることなく、被写体を画像に収めることができるVizPhotoを提案しています。VizPhotoは、記録された音声情報と撮影日時に加え、物体検出により検出された物体のリストを用いて、Webインタフェース上で全方位画像を選択し、検出された物体から任意の物体を選択し、透視投影画像を出力する。この方法は、画角内の対象物を撮影することには成功しているが、自撮りを目的としたものではなく、人間の表情を考慮したものではない。

Unlike the previous studies, this study focuses on taking selfies and aims to output a perspective projection image in which all the faces are in the angle of view and have good facial expressions.
%%%先生の英語の和訳
%本研究では、従来の研究とは異なり、自撮りに着目し、すべての顔が画角内にあり、表情の良い透視投影画像の出力を目指す。

\section{Proposed Method}
\subsection{Outline}
%Figure \ref{fig1} shows the processing procedure of the proposed method. 
In the proposed method, first, a user takes a few seconds of omni-directional video. Second, the method performs face detection for all persons in all frames in the omni-directional video. Third, the method eliminates false detection and interpolates undetected faces. Fourth, the method performs facial expression recognition on all the frames, and extract the frame with the highest happiness value for all the participants. Finally, the method transforms the omni-directional image to the perspective projection image. In this step, it calculates the image center and the angle of view so that all the faces are within the frame. In the following, we describe each step in details.

%フローやパワーポイントはともかく，英語で主語のない文は絶対にいけません．
% the というのは，文を読んだ人が，あぁそのことね，とわかるものの前につけるものなので，the shooting conditionsと書いてるといったいなんのことなのか，読み手は？？？になります．
%The processing procedure of the proposed method is shown in Fig. \ref{fig1}. First, a few seconds of omni-directional video is taken according to the shooting conditions. Second, performs face detection for all persons in the omni-directional video for all frames. Third, eliminates false detection except for faces and interpolates the coordinates of undetected faces. Fourth, facial expression recognition is performed on all frames. Fifth, the frame with the highest overall happiness value is extracted. Finally, transform the omni-directional image to the perspective projection image. If there are multiple faces in the frame, extracts the coordinates of the leftmost and rightmost faces. And, Calculate the angle of view so that all the faces are within the frame centered at the average position of the edges of the face areas of these persons.

%全ての項目で書き方（名詞形なのか動詞で始まる文なのか）を統一すること
%\begin{figure}
%\begin{center}
%\includegraphics[width=0.8\textwidth]{figs/1.png}
%\caption{Flow of the proposed method.} 
%\label{fig1}
%\end{center}
%\end{figure}

\subsection{Omni-directional Video Capture}
In the proposed method, a user takes a selfie with an omni-directional camera according to the following three conditions.
%In the proposed method, the following three conditions are set for taking a selfie with an omni-directional camera.
\begin{enumerate}
    \item A user extends his/her arm and holds the camera at about the same height as his/her shoulder.
    \item Persons to be photographed are within about 1.5 meters from the camera.
    \item Persons to be photographed gather in one direction from the camera and do not spread out to the sides.
\end{enumerate}
This study also assumes that the downward direction of the omni-directional image is the direction of gravity in the real world, using the direction of gravity obtained by the acceleration sensor inside the omni-directional camera.
%\begin{enumerate}
%    \item Extend your arm and position the camera at the same height as your shoulder.
%    \item Subject of a photograph must be within about 1.5 meters of the camera
%    \item Angle of view does not exceed approximately 180°
%\end{enumerate}

\subsection{Frame Extraction with Face Detection and Expression Recognition}
\subsubsection{Face Detection on All Frames}
\begin{figure}[t]
\begin{center}
\begin{minipage}{0.16\textwidth}
\includegraphics[width=\textwidth]{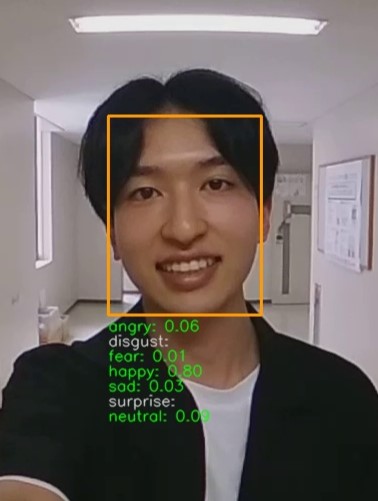}
\subcaption{Face}
\label{fig3a}
\end{minipage}
\begin{minipage}{0.18\textwidth}
\mbox{\raisebox{-26mm}{\includegraphics[width=\textwidth]{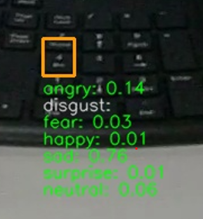}}}
\subcaption{Keyboard}
\label{fig3b}
\end{minipage}
\begin{minipage}{0.18\textwidth}
\mbox{\raisebox{-26mm}{\includegraphics[width=\textwidth]{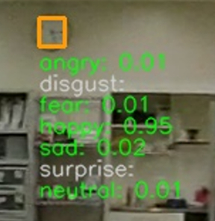}}}
\subcaption{Clock}
\label{fig3c}
\end{minipage}
\begin{minipage}{0.15\textwidth}
\includegraphics[width=\textwidth]{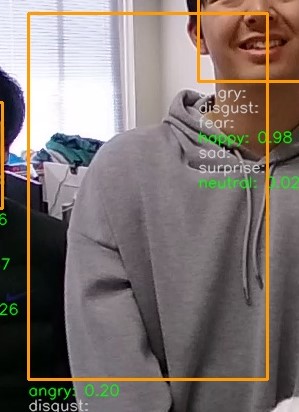}
\subcaption{Nothing}
\label{fig3d}
\end{minipage}
\caption{Example of detections}
\label{fig3}
\end{center}
\end{figure}

\begin{figure}[t]
\begin{center}
\includegraphics[width=0.4\textwidth]{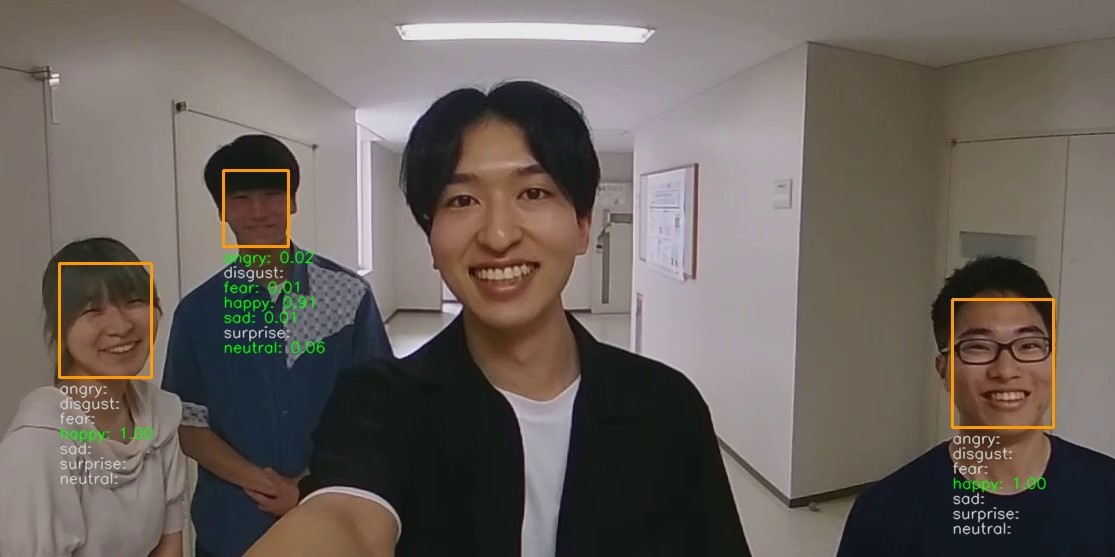}
\caption{Example of non-detection.}
\label{fig4}
\end{center}
\end{figure}

The proposed method detects faces from all frames in an omni-directional video. Here, We use MTCNN\cite{zhang2016joint} (Multi-task Cascaded Convolutional Neural Networks). Figure \ref{fig3}(a) shows an example of face detection, in which the face detection is successful. However, objects other than faces such as keyboards and clocks can also be falsely detected as shown in Figs. \ref{fig3}(b)(c)(d). In addition, MTCNN often fails to detect faces as shown in Fig. \ref{fig4}. This may be caused by various factors such as brightness, hair, and hats. 

To reduce the false detection, assuming that persons to be photographed are within approximately 0.3 to 1.5 meters from the camera as mentioned above, we set the minimum lengths for one side of the bounding box. This reduces the number of false detections such as the small bounding boxes shown in Figs. \ref{fig3}(b)(c). However, since this alone does not completely prevent false detections, further methods to prevent false detections are necessary.

%The proposed method detects faces from an omni-directional image because all the photographers must fit in the angle of view. We use MTCNN\cite{MTCNN} (multi-cascade convolutional network) of detecte faces in an image using three convolutional networks for face detection. Figure 2 shows an example of face detection. In this example, the face detection is successful and the bounding box is obtained. However, false detections other than faces may occur in some of the frames. Figure 3 shows examples of false detections such as chair, keyboard, clock, and monitor when face detection is performed. In addition, MTCNN often identifies undetected faces as shown in Figure 4. This is caused by various factors such as brightness, hair, and hats. Therefore, the minimum face size setting for faces detect by MTCNN is changed from the default of 20 pixel squares to 50 pixel squares for reduce false detections and undetected faces. This reduces the number of false detections represented by the small bounding box shown in Figure 3 and satisfies shooting condition 2. In addition, it is designed to prevent detection of unrelated persons who are approximately 1.5 meters to 3 meters away from the camera. However, this is a reduction of false detections and undetections due to changes in the MTCNN settings, and further reduction methods are required.

%\begin{figure}[tbh]
%\begin{center}
%\includegraphics[width=0.25\textwidth]{figs/2.jpg}
%\caption{Example of Face Detection} 
%\label{fig2}
%\end{center}
%\end{figure}

\subsubsection{Eliminating False Detection Faces and Interpolating Undetected Faces}
To further reduce the reduction of false detections and also interpolate undetected faces, we first calculate the center coordinates of the bounding boxes from the upper left coordinates, width and height of bounding boxes for all the detected faces in all the frames. Second, we cluster all the center coordinates of bounding boxes using Mean Shift method \cite{comaniciu2002mean}. Third, if the ratio of the number of elements in a class to the total number of frames is below a certain threshold $T$, the bounding boxes in the class are deleted as false detection. In addition, we set the maximum length of one side of the bounding box and if the ratio of the number of elements in a class that have large regions as shown in Fig. \ref{fig3}(d) is more than $T$, we eliminate the class. Finally, if an element in a remaining class is not present in a certain frame, a bounding box is generated in that frame using the mean coordinates and the mean width and height of the bounding boxes in the class. This allows undetected faces to be interpolated. The face detection results with false detection eliminated and undetected faces interpolated are used for the next facial expression recognition.

\subsubsection{Facial Expression Recognition and Extraction of a Frame}
The proposed method applies FER (Face Emotion Recognizer)\cite{justin_shenk_2021_5362356}, which recognizes the facial expressions in seven categories: angry, disgust, fear, happy, sad, surprise and neutral. Based on the recognition results, the frame with the highest value is extracted from the omni-directional video using Eq. (\ref{eq:1}).
\begin{equation}
H = M - cV,
\label{eq:1}
\end{equation}
where $M$ and $V$ represent the mean and variance of the happiness expression values of all the detected faces in a frame and $c$ is a coefficient. The overall happiness value $H$ defined in Eq. (\ref{eq:1}) is calculated for each frame and the frame with the highest $H$ is extracted. In this study, we define happiness as the mean minus the variance, based on the idea that it would be ideal if everyone's facial expressions were as uniform as possible.
%本研究では，できるだけ全員の表情がばらついていない方が理想的であるという考えに基づき，平均から分散を引いた値を幸せ度として定義する．

\begin{figure}[tb]
\begin{center}
\includegraphics[width=0.9\textwidth]{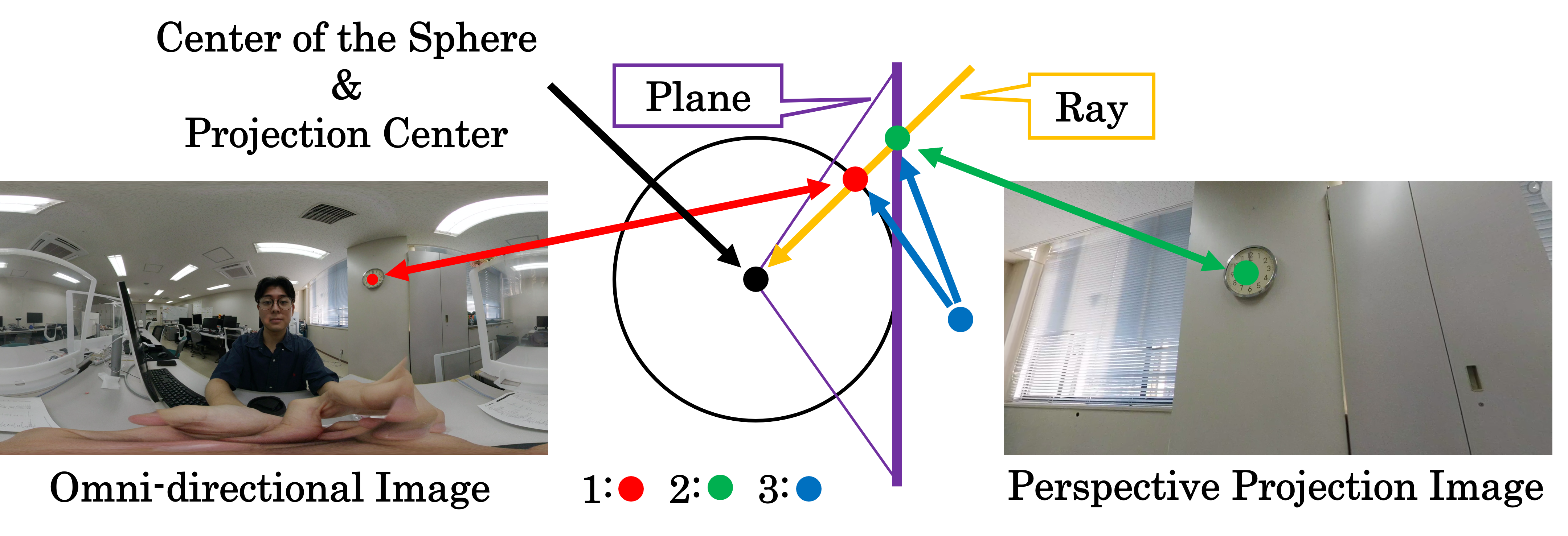}
\caption{Relationship between omni-directional and perspective projection images.}
\label{fig6}
\end{center}
\end{figure}

\subsection{Transformation to Perspective Projection Image}
The proposed method transforms the omni-directional frame with the highest $H$ to a perspective projection image in which all the participants are in the angle of view. Figure \ref{fig6} shows the relationships for the transformation. The transformation from an omni-directional image to a perspective projection image is obtained by considering the following three relationships 1, 2, and 3.
\begin{enumerate}
    \item A coordinate in an omni-directional image and a 3D position of a sphere
    \item A coordinate in a perspective projection image and a 3D position of a plane
    \item 3D positions of a sphere and a plane through which the same ray passes
\end{enumerate}
In addition, according to the positions of the photographed faces, the proposed method determines the center of the perspective projection image and the angle of view. We describe each relationship in detail below.

\subsubsection{Relationship between Coordinate in Omni-directional Image and 3D Position of Sphere}
%Figure \ref{fig7} shows the 3D position of the sphere. The 3D coordinates $(x, y, z)$ can be expressed using two angles $(\theta, \varphi)$ and a radius $r$ as follows:
%\begin{equation}
%\left\{
%\begin{aligned}
%& x = r\sin\theta \cos\varphi \\
%& y = r\cos\theta \sin\varphi \\
%& z = r\cos\theta .
%\label{eq:2}
%\end{aligned}
%\right.
%\end{equation}
%Assuming that the color of the rays is not attenuated, we can set $r=1$ because the relationship between the omni-directional image and the sphere no longer depends on $r$.

%\begin{figure}[tb]
%\begin{center}
%\begin{minipage}{0.44\textwidth}
%\includegraphics[width=\textwidth]{figs/7.png}
%\caption{3D position of the sphere} 
%\label{fig7}
%\end{minipage}
%\begin{minipage}{0.55\textwidth}
%\mbox{\raisebox{-32mm}{\includegraphics[width=\textwidth]{figs/8.png}}}
%\caption{Coordinates in omni-directional image} 
%\label{fig8}
%\end{minipage}
%\end{center}
%\end{figure}

Coordinates $(u,v)$ in an omni-directional image with equirectangular projection are corresponding to $\phi$ and $\theta$, which are angles around and from Z-axis connecting the top and bottom of the sphere, respectively.
Given that $h_o$ and $w_o$ denote the vertical and horizontal lengths of the omni-directional image, respectively, coordinates $(u,v)$ of the omni-directional image can be expressed using angles $\phi$ and $\theta$ as follows:
%Figure \ref{fig8} shows an omni-directional image using the equirectangular projection method. In the equirectangular projection image, the $x$ and $y$ coordinates correspond to the angles $\phi$ and $\theta$ in Fig. \ref{fig7}. Therefore, given that $h_o$ and $w_o$ denote the vertical and horizontal lengths of the omni-directional image, respectively, angles $\phi$ and $\theta$ can be expressed using the coordinates $(u,v)$ of the omni-directional image as follows:
\begin{equation}
\left\{ \,
    \begin{aligned}
    & u=\frac{w_o}{2\pi\varphi} \\
    & v=\frac{h_o}{\pi\theta} .
    \end{aligned}
\right.
\label{eq:3}
\end{equation}

%From Equations (3), (6) and (7), the relationship between the coordinates of the omni-directional image and the 3D position of the sphere can be expressed by Equation (8).
%\begin{equation}
%\left\{
%\begin{aligned}
%x = \sin\frac{\pi v}{h_p} \cos\frac{2\pi u}{w_p} \\
%y = \cos\frac{\pi v}{h_p} \sin\frac{2\pi u}{w_p} \\
%z = \cos\frac{\pi v}{h_p} 
%\end{aligned}
%\right.
%\end{equation}

\subsubsection{Relationship between Coordinate in Perspective Projection Image and 3D Position of Plane}
\begin{figure}[tb]
\begin{center}
\begin{minipage}{0.28\textwidth}
\includegraphics[width=\textwidth]{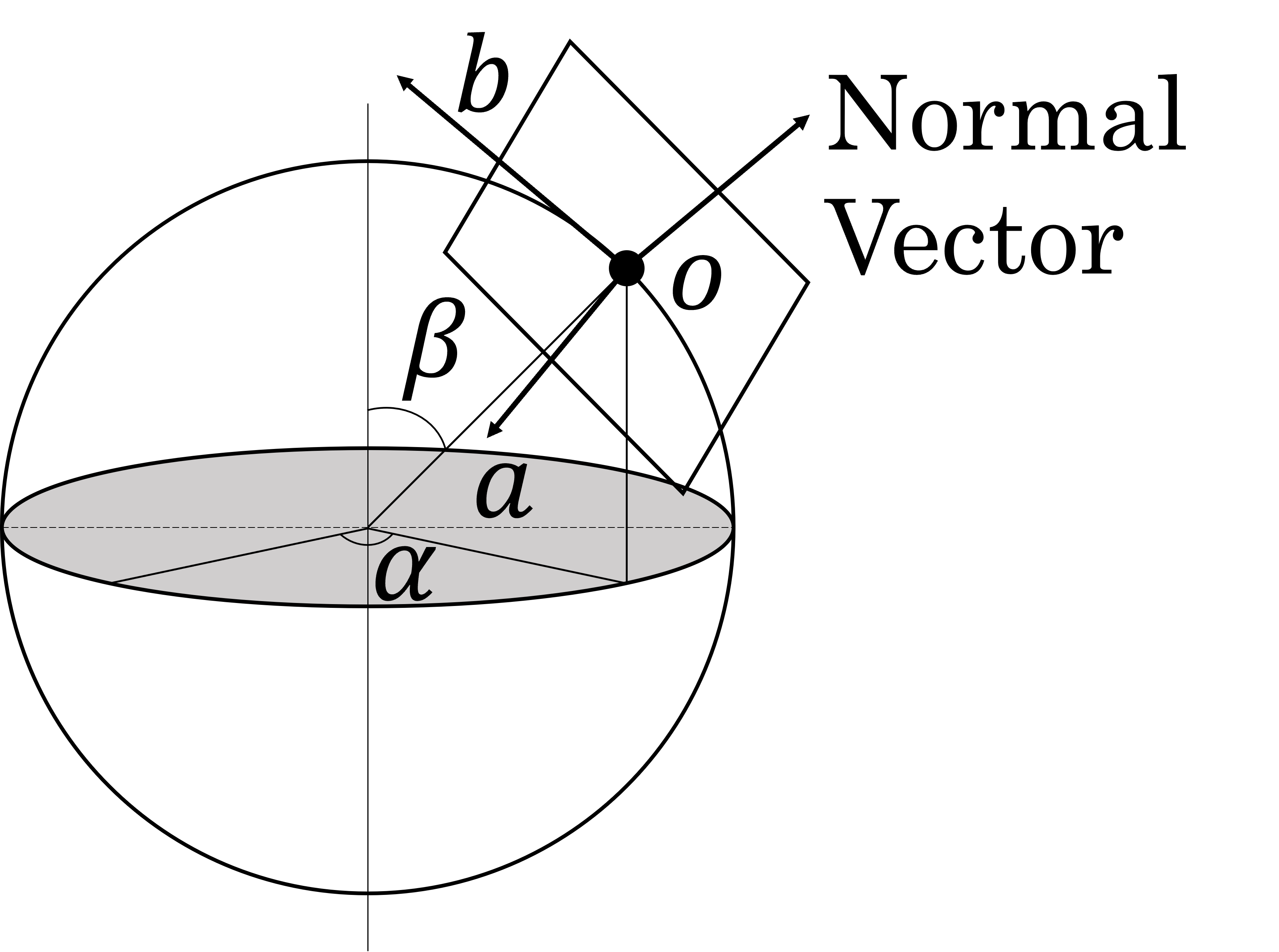}
\caption{3D position of plane on sphere}
\label{fig9}
\end{minipage}\hspace{4mm}
\begin{minipage}{0.62\textwidth}
\includegraphics[width=\textwidth]{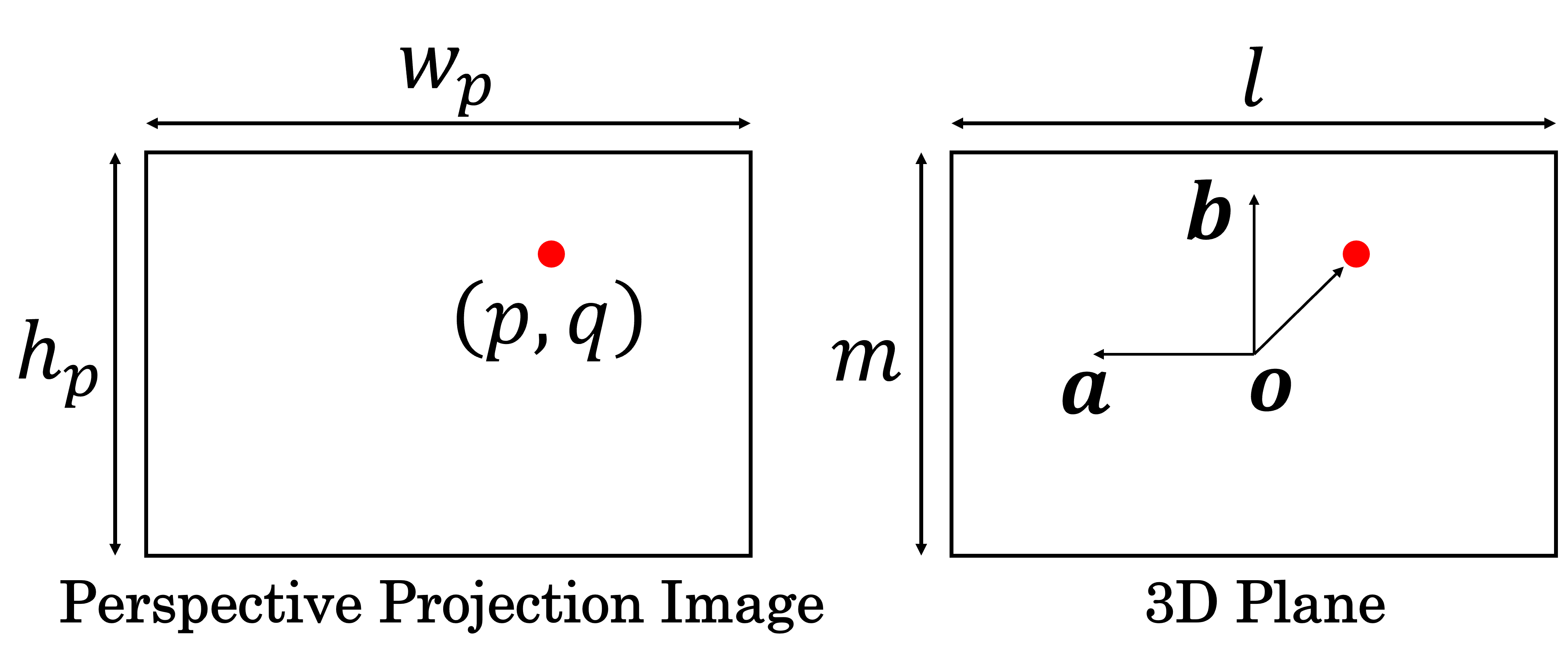}
\caption{Relationship between perspective projection image and 3D position of plane.}
\label{fig10}
\end{minipage}
\end{center}
\end{figure}

Figure \ref{fig9} shows the 3D position of the plane, which is corresponding to a perspective projection image, tangent to the sphere. The plane consists of the center position $\boldsymbol{o}$, and two orthogonal unit vectors $\boldsymbol{a}$ and $\boldsymbol{b}$. They respectively become the center of the perspective projection image, and horizontal and vertical axes. In this study, $\boldsymbol{o}$, $\boldsymbol{a}$ and $\boldsymbol{b}$ are defined so that the upward direction of the image becomes that of the real world as follows:
\begin{equation}
\boldsymbol{o}=
\begin{pmatrix}
\sin\beta \cos\alpha \\
\sin\beta \sin\alpha \\
\cos\beta 
\end{pmatrix},
\boldsymbol{a}=
\begin{pmatrix}
\sin\alpha \\
 -\cos\alpha \\
0
\end{pmatrix}, 
\boldsymbol{b}=
\begin{pmatrix}
-\cos\alpha \cos\beta \\
-\sin\alpha \cos\beta \\
\sin\beta
\end{pmatrix},
\end{equation}
where $\alpha$ and $\beta$ are angles around and from Z-axis connecting the top and bottom of the sphere, respectively.

Figure \ref{fig10} shows the relationship between a coordinate in a perspective projection image and the 3D position on the plane. The left and right figures in Fig. \ref{fig10} represents a perspective projection image with height $h_p$ and width $w_p$, and a 3D plane with height $m$ and width $l$, respectively. In this case, the 3D position on the plane is represented using parameter $(s,t)$ as follows:
\begin{equation}
\boldsymbol{o}+s\boldsymbol{a}+t\boldsymbol{b}.
\label{eq:4}
\end{equation}
From Fig. \ref{fig10} and Eq.(\ref{eq:4}), the relationship between the coordinate $(p,q)$ in the perspective projection image and $(s,t)$ is expressed by a translation and scaling matrix as follows.
\begin{equation}
\begin{bmatrix}
s \\
t \\
1 
\end{bmatrix}
=
\begin{bmatrix}
s_x & 0 & t_x \\
0 & s_y & t_y \\
0 & 0 & 1
\end{bmatrix}
\begin{bmatrix}
p \\
q \\
1
\end{bmatrix}.
\label{eq:5}
\end{equation}

\begin{figure}[t]
\begin{center}
\includegraphics[width=0.33\textwidth]{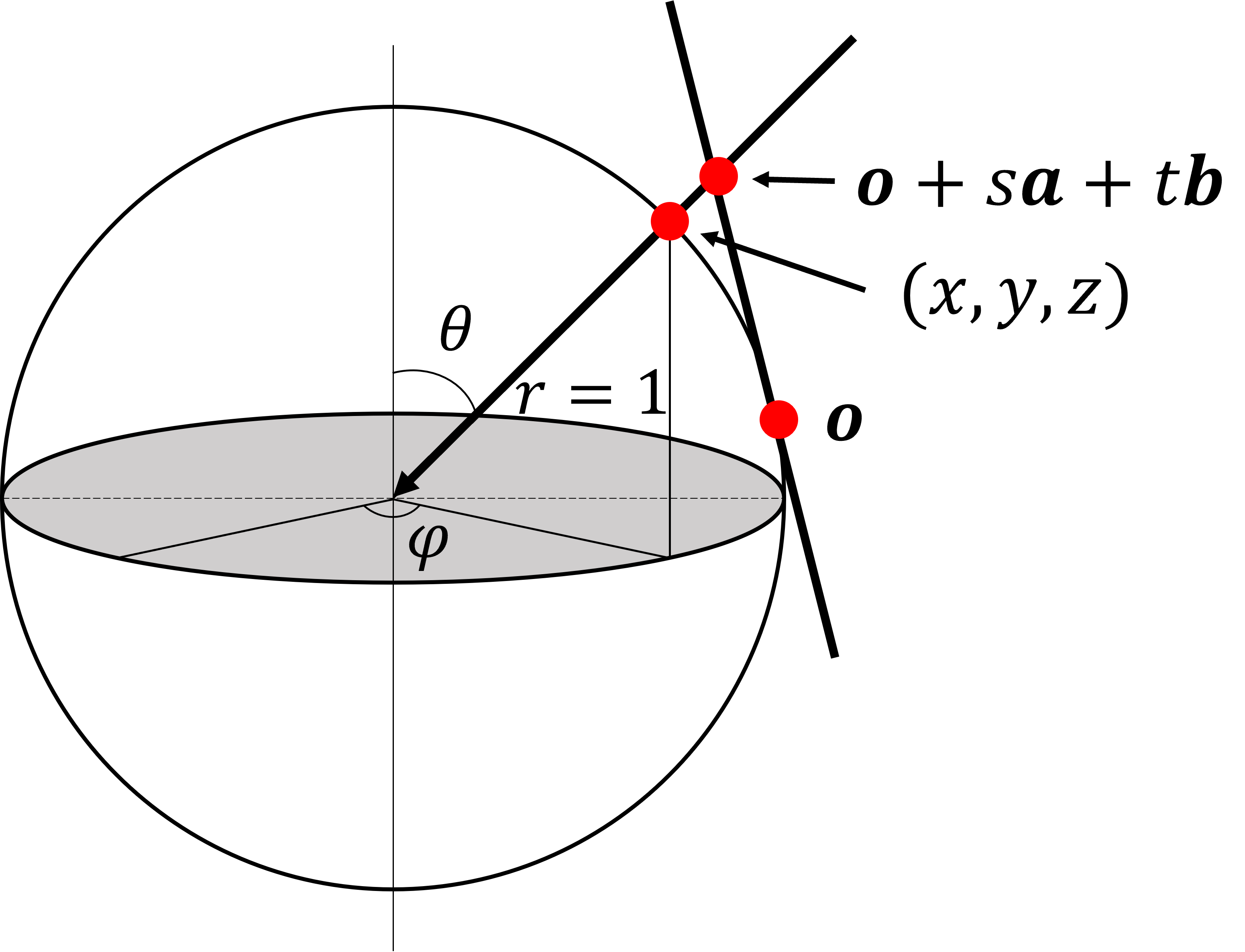}
\caption{Relationship between the 3D positions of a sphere and a plane through which the same ray passes}
\label{fig11}
\end{center}
\end{figure}

\subsubsection{Relationship between 3D Positions of Sphere and Plane through which the Same Ray Passes}
Figure \ref{fig11} shows the relationship between points on the sphere and the plane where the same ray passes. Here, since the line passing through the point on the sphere and the center of the sphere intersects the plane, the relationship can be expressed with parameter $k$ as follows:
\begin{equation}
k
\begin{bmatrix}
\sin\theta \cos\varphi \\
\sin\theta \sin\varphi \\
\cos\theta
\end{bmatrix}
=\boldsymbol{o}+s\boldsymbol{a}+t\boldsymbol{b}.
\label{eq:8}
\end{equation}

From the above Eqs. (\ref{eq:3}), (\ref{eq:5}) and (\ref{eq:8}), the coordinates $(p,q)$ in the perspective projection image can be mapped to the coordinates $(u,v)$ in the omnidirectional image to generate the perspective projection image. Here, it is necessary to find the center $\boldsymbol{o}$ of the plane and the lengths $l$ and $m$ of the edges of the plane, which are corresponding to the center and the view angle of the perspective projection image so that the all the participants photographed are within the image.

\subsubsection{Determination of the most left, right, top and bottom faces}
This study deals with omni-directional images in which the left and right edges of the images are connected. For this reason, the left and right positions of the faces in the omni-directional image do not necessarily correspond to their actual positions. Therefore, the left-most and right-most faces that appear in the perspective projection image are determined. 

\begin{figure}[t]
\begin{center}
\includegraphics[width=0.35\textwidth]{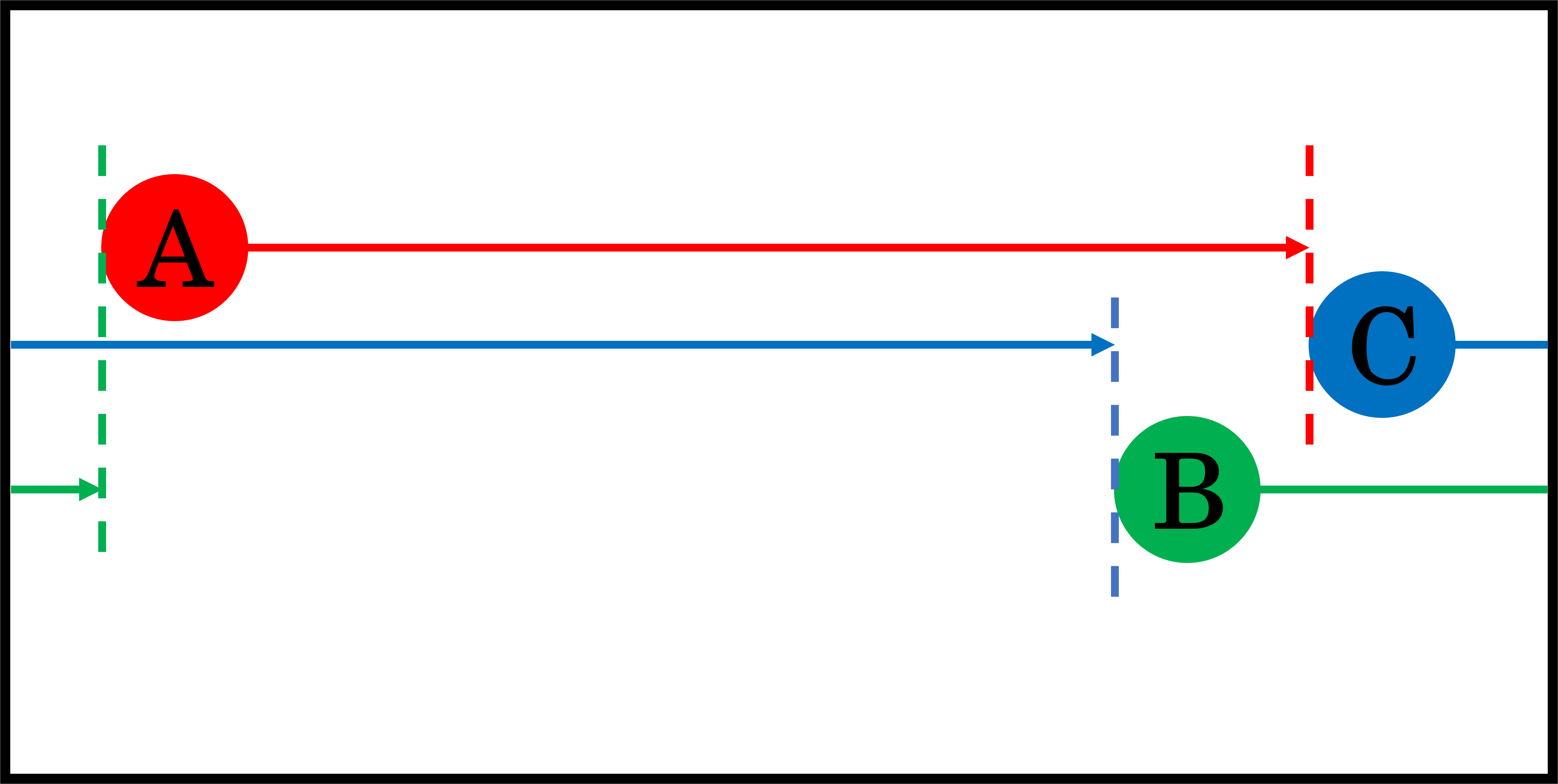}
\caption{Determination of the most left and right faces.}
\label{fig12}
\end{center}
\end{figure}

The proposed method calculates the distance from a face to the face with the longest distance from it in the positive direction of the x-axis of the omni-directional image. This is done for all faces, and the pair of faces with the shortest distance is extracted as the left face and the right face. Figure \ref{fig12} illustrates an example in which A, B, and C represent three faces. First, if we consider A to be the left-most face and look for the face farthest to the left in the positive direction of the x-axis, we find C and calculate the distance between A and C. The same process is performed when B and C are considered the left-most faces. Next, the method finds the shortest distance. In the case of Fig. \ref{fig12}, B and A are the leftmost and the rightmost faces, respectively. As for the most top and bottom faces, we simply find the faces with the smallest and largest y-coordinates.

\subsubsection{Calculation of Center on Perspective Projection Image}
The proposed method calculates 3D coordinate $\boldsymbol{o}$ on the plane, which is corresponding to the center of the perspective projection image. Specifically, we obtain the mean of the x-coordinates of the bounding box centers of the most left and right faces, and the mean of the y-coordinates of the bounding box centers of the most top and bottom faces. Note that the average of x is calculated considering only in the positive direction of the x-axis. Using the equations mentioned above, we transform x and y coordinates into angles $(\alpha,\beta)$ to obtain 3D position $\boldsymbol{o}$.

\subsubsection{Calculation of Angle of View of Perspective Projection Image}
Since the vertical and horizontal lengths $l$ and $m$ of the 3D plane shown in Fig. \ref{fig10} determine the 
angle of view of the perspective projection image, $(l,m)$ is calculated from the coordinates of the faces.
Specifically, $s$ and $t$ are obtained from the coordinates of the corners of all bounding boxes using Eqs (\ref{eq:3}) and (\ref{eq:8}), and the maximum absolute values of $s$ and $t$ are obtained.
%具体的には，全てのバウンディングボックスのコーナーの座標から式(2)および式(6)によりsとtを求め，sとtの絶対値の最大値を求める．
Since $\boldsymbol{a}$ and $\boldsymbol{b}$ are unit vectors, doubling $s$ and $t$ yields an angle of view that includes all the faces. However, as shown in Fig. \ref{fig3}(a), the bounding box does not include the hair area. Therefore, by adding offset $i$, $l$ and $m$ are determined as follows:
\begin{equation}
    \left\{ \,
    \begin{aligned}
    & l=2s+i \\
    & m=2t+i. 
    \end{aligned}
\right.
\label{eq:11}
\end{equation}
%$s$ and $t$ represent the $x$ coordinate of the leftmost face and the $y$ coordinate of the topmost face to the center of the perspective projection image in the $x$ axis and $y$ axis directions in the 3-D plane, respectively. However, the face detection example in Fig. \ref{fig2} shows that hair and other parts of the face may not be included in the bounding box coordinates, and the face may be slightly obscured. Therefore, $(l,m)$ is calculated by adding the constant $i$ so that all the faces are within the angle of view. Finally, the number of pixels in the horizontal and vertical directions of the perspective projection image is calculated by substituting $(l,m)$ into Equations (18) and (19) and multiplying by the factor $j$. 
%\begin{equation}
%    \left\{ \,
%    \begin{aligned}
%    & w=lj \\
%    & h=mj 
%    \end{aligned}
%\right.
%\label{eq:12}
%\end{equation}

\section{Experiments and Discussions}
To demonstrate the effectiveness of the proposed method, we performed comparison experiments with a baseline method. For the baseline method, we used the default settings of MTCNN, i.e., the minimum value of one side of the bounding box for face detection is 20 pixels and the maximum value is not fixed. To extract the frame with the highest $H$ from all frames, the baseline method did not eliminate false detections or undetections, but simply extracted the frames in which the number of faces detected in a frame was the mode throughout all frames as a mechanism for eliminating false detections and undetections instead, and extracted the frame with the highest $H$ from the frames.

For the experiments, we used five types of omni-directional videos ($3840 \times 1920$ pixels) with different conditions, as shown in Table \ref{ExConditions1}, and performed three types of validation. Experiment 1 is for validation of happy values, Experiments 2 and 3 are for validation of false detections, and Experiments 4 and 5 are for validation of undetections. These experiments were conducted by sighted people, not visually impaired people, and we empirically set the parameters of the proposed method as shown in Table \ref{ExConditions2}.
%提案手法の有用性を示すために我々はベースライン手法との比較実験を行った．また，表1と2に示す6種類の全方位動画および条件により透視投影画像の生成を行い，3種類の検証を行った．実験1は幸せ値に関する検証，実験2・3・4は誤検出に関する検証，実験5・6は未検出に関する検証である．今回の実験では視覚障害者ではなく晴眼者によって実験を行った．幸せ値と(s,t)と解像度に関する実験結果を表3，提案手法の検出に関する実験結果を表4，ベースライン手法の検出に関する実験結果を表5に示す．以下では，3種類の検証について順に述べる
%To demonstrate the usefulness of the proposed method, we performed a comparison experiment with the baseline method. Also, the perspective projection images were generated by the 7 omni-directional videos and conditions shown in Tabs. \ref{ExConditions1} and \ref{ExConditions2} , and 3 types of verification were performed. Experiments 1 is verification of happy values, Experiments 2, 3 and 4 are verification of false detections, and Experiments 5 and 6 verification of undetected values. In this experiment, these experiments were conducted by sighted people, not visually impaired people. 

Table \ref{ExResults} shows the experimental results of $H$ and $(s, t)$ of perspective projection images. Table \ref{proposed_detection} shows the experimental results for detection by the proposed method, and Table \ref{baseline_detection} shows the experimental results for detection by the baseline method. In the following, we discuss the three types of validation in turn.

\begin{table}[tbhp]
\caption{Experimental conditions.}\label{ExConditions1}
\centering
\begin{tabular}{|c|c|c|}
\hline
\textbf{Expt.} & \textbf{No. of participants} & \textbf{No. of unrelated persons}   \\
\hline\hline
1 & 6 & 5 \\  
2 & 3 & 0 \\ 
3 & 2 & 1 \\ 
4 & 5 & 0 \\ 
5 & 7 & 4 \\ 
\hline
\end{tabular}
%\end{table}
%\begin{table}[b]
\caption{Parameters of the proposed method.}\label{ExConditions2}
\centering
\begin{tabular}{|c|c|}
\hline
Minumum length of bounding box & 50 pixels \\ \hline
Maximum length of bounding box & 250 pixels \\ \hline
Threshold $T$ for eliminating false detections & 25 \% \\ \hline
Coefficient $c$ for happy value & 1 \\ \hline
Offset $i$ for perspective projection image & 0.3 \\
\hline
\end{tabular}
%\end{table}
%\begin{table}[tb]
\caption{Experimental results of the proposed and baseline methods for $H$, $(s, t)$.}\label{ExResults}
\centering
\begin{tabular}{|c||c|c||c|c|}
\hline
& \multicolumn{2}{|c||}{Proposed} & \multicolumn{2}{|c|}{Baseline} \\ \hline
\textbf{Expt.} &\textbf{$H$} & \textbf{$(s, t)$} &  \textbf{$H$} & \textbf{$(s, t)$} \\
\hline
1 & 0.458 & $(1.880, 0.454)$ & 0.400 & $(1.883, 0.452)$  \\
2 & 0.895 & $(1.213, 0.469)$ &  0.648 & $(1.959, 0.622)$  \\
3 & 0.309 & $(0.335, 0.216)$ &  0.155 &  $(11.82, 1.677)$  \\
4 & 0.483 & $(1.057, 0.363)$ &  0.612 & $(1.071, 0.362)$  \\
5 & 0.402 & $(2.294, 0.504)$ &  0.491 &  $(145.4, 34.72)$  \\
\hline
\end{tabular}
%\end{table}
%\begin{table}[tb]
\caption{Experimental results of the proposed method for detection. (a) No. of frames, (b) No. of classes after eliminating false detection, (c) No. of classes by Mean-shift, (d) No. of frames in which false detection occurred before eliminating false detection, and (e) No. of frames in which undetected participants existed before interpolation.}\label{proposed_detection}
\centering
\begin{tabular}{|c|c|c|c|c|c|}
\hline
\textbf{Expt.} & (a) & (b) & (c) & (d) & (e) \\ \hline
1 & 51 & 6 & 7 & 1 & 32 \\
2 & 143 & 3 & 7 & 32 & 10 \\
3 & 152 & 2 & 2 & 0 & 0 \\
4 & 41 & 5 & 5 & 0 & 32 \\
5 & 35 & 7 & 8 & 1 & 30 \\
\hline
\end{tabular}
%\end{table}
%\begin{table}[tb]
\caption{Experimental results of the baseline method for detection. (a) No. of frames, (b) Mode throughout all frames, (c) No. of frames whose number of detections is mode, (d) Ratio of frames with No. of detections is mode, (e) No. of frames in which false detection occured, (f) No. of frames in which undetected participants existed, (g) No. of ideal frames in which no false detections and no undetections, and (h) Ratio of ideal frames.}\label{baseline_detection}
\centering
\begin{tabular}{|c|c|c|c|c|c|c|c|c|}
\hline
\textbf{Expt.} & (a) & (b) & (c) & (d) & (e) & (f) & (g) & (h) \\
\hline
1 & 51 & 6 & 19 & 0.372 & 31 & 38 & 5 & 0.098 \\
2 & 143 & 4 & 52 & 0.363 & 101 & 2 & 41 & 0.286 \\
3 & 152 & 3 & 108 & 0.710 & 147 & 2 & 5 & 0.032 \\
4 & 41 & 4 & 25 & 0.609 & 1 & 39 & 2 & 0.048 \\
5 & 35 & 6 & 13 & 0.371 & 34 & 32 & 0 & 0.000 \\
\hline
\end{tabular}
\end{table}

\subsection{Validation on Happiness Values}
\begin{figure}[t]
\begin{center}
\begin{minipage}{0.32\textwidth}
\includegraphics[width=\textwidth]{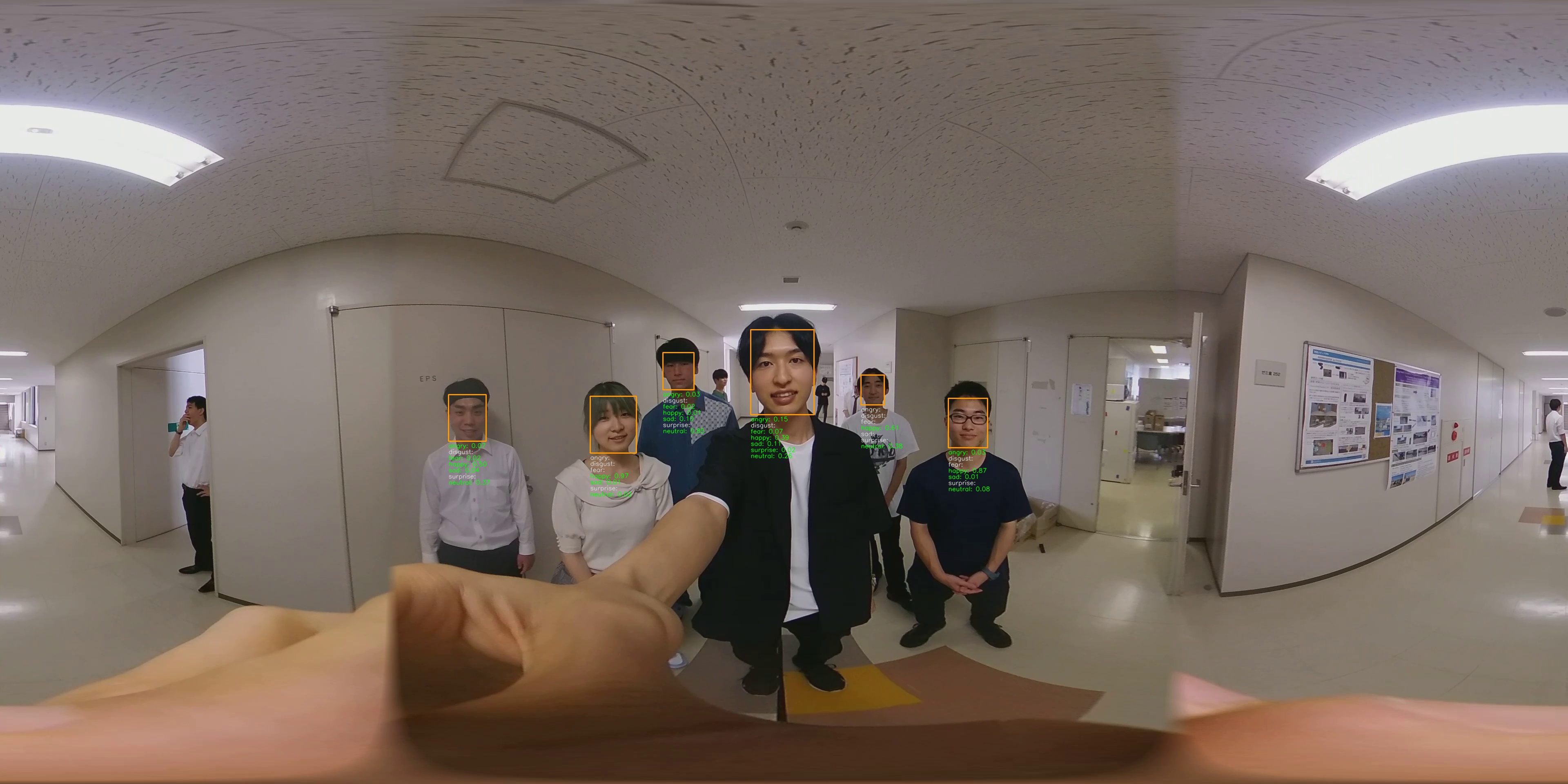}
\subcaption{Frame with highest $H$ by the proposed method} 
\label{Ex1pF}
\end{minipage}
\begin{minipage}{0.32\textwidth}
\includegraphics[width=\textwidth]{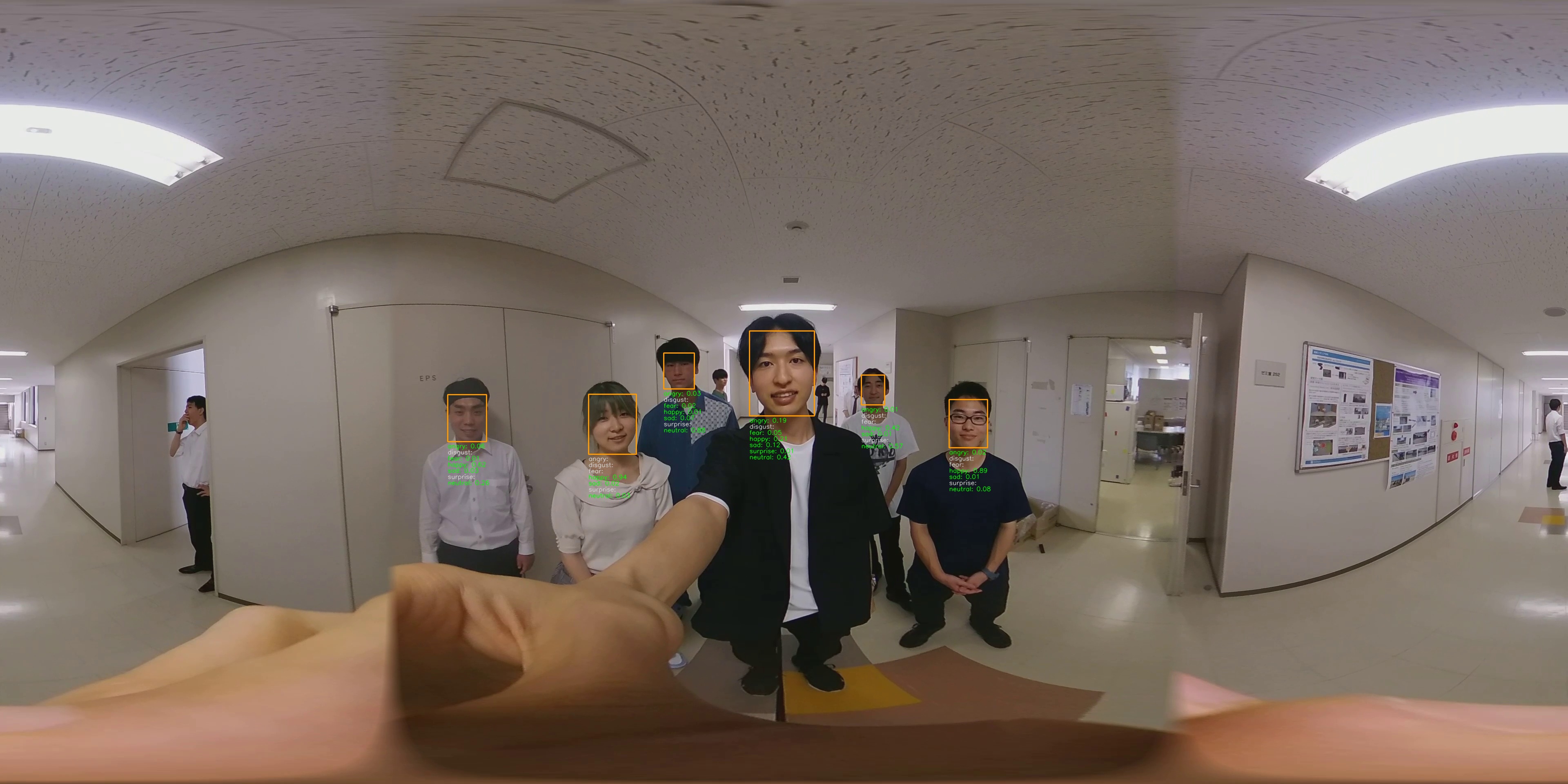}
\subcaption{Frame with highest $H$ by the baseline method} 
\label{Ex1bF}
\end{minipage}
\begin{minipage}{0.32\textwidth}
\includegraphics[width=\textwidth]{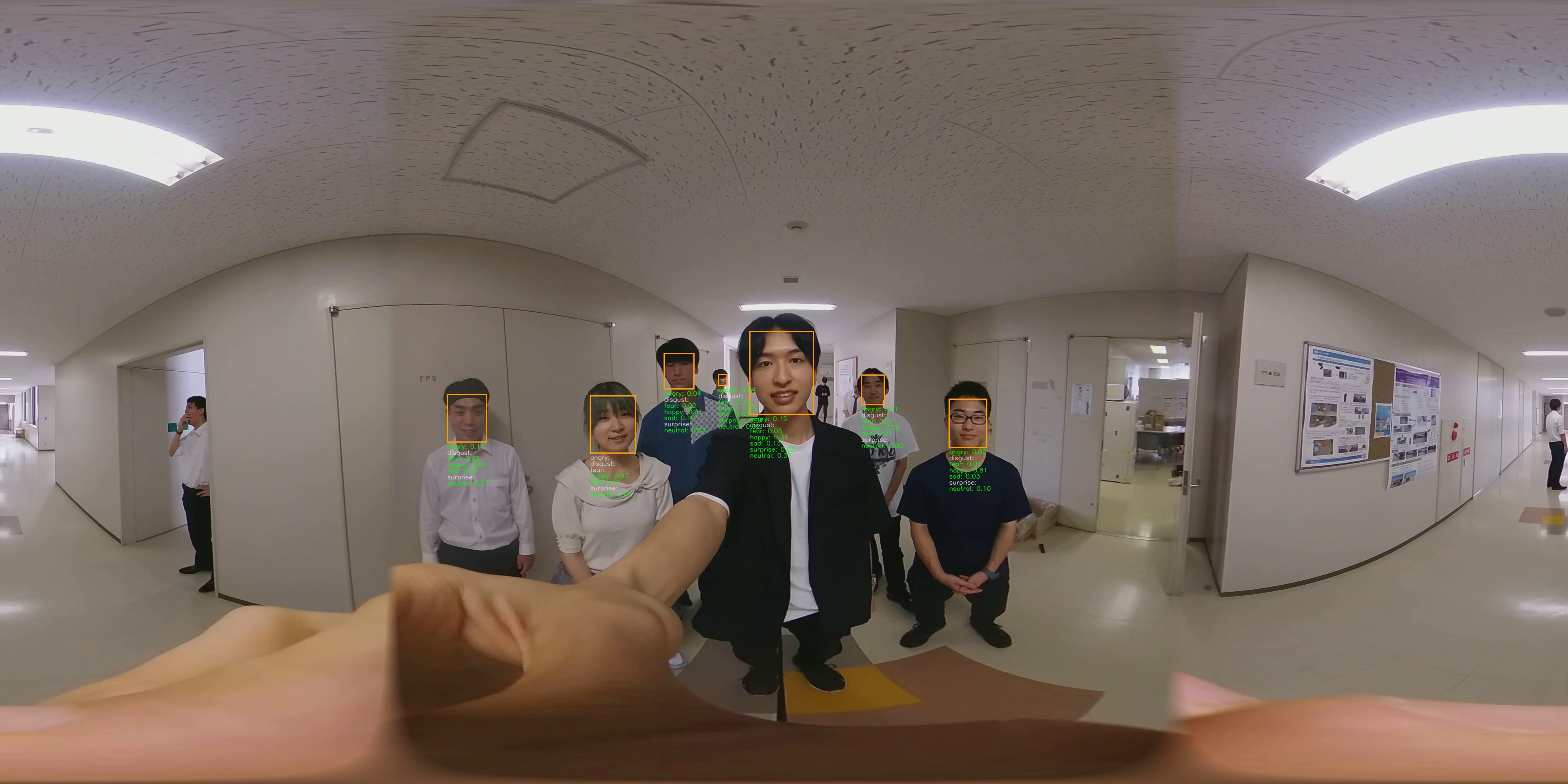}
\subcaption{Same frame as (a) by the baseline method} 
\label{Ex1bF2}
\end{minipage}
\begin{minipage}{0.35\textwidth}
\includegraphics[width=\textwidth]{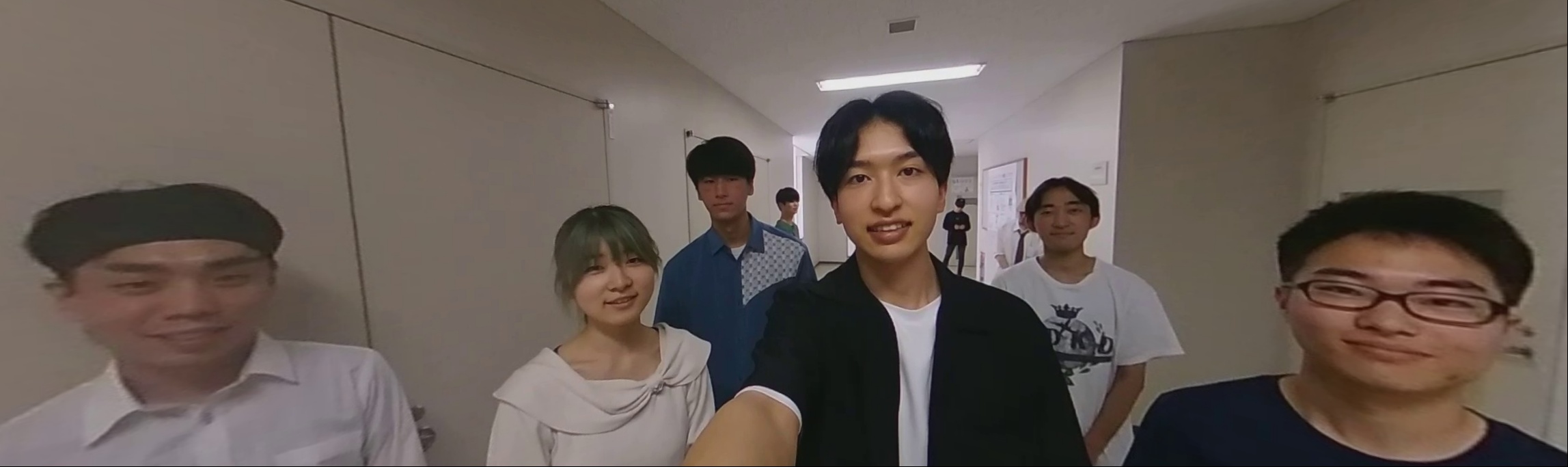}
\subcaption{Generated image from (a)}
\label{Ex1pp}
\end{minipage}
\begin{minipage}{0.35\textwidth}
\includegraphics[width=\textwidth]{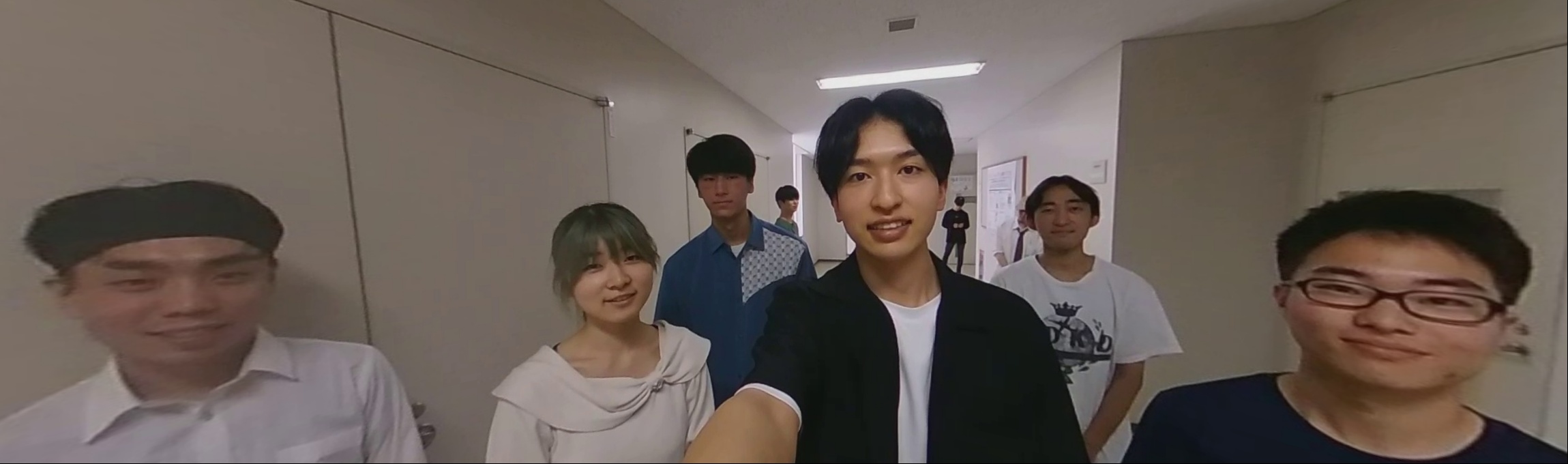}
\subcaption{Generated image from (b)}
\label{Ex1bp}
\end{minipage}
\begin{minipage}{0.28\textwidth}
\includegraphics[width=\textwidth]{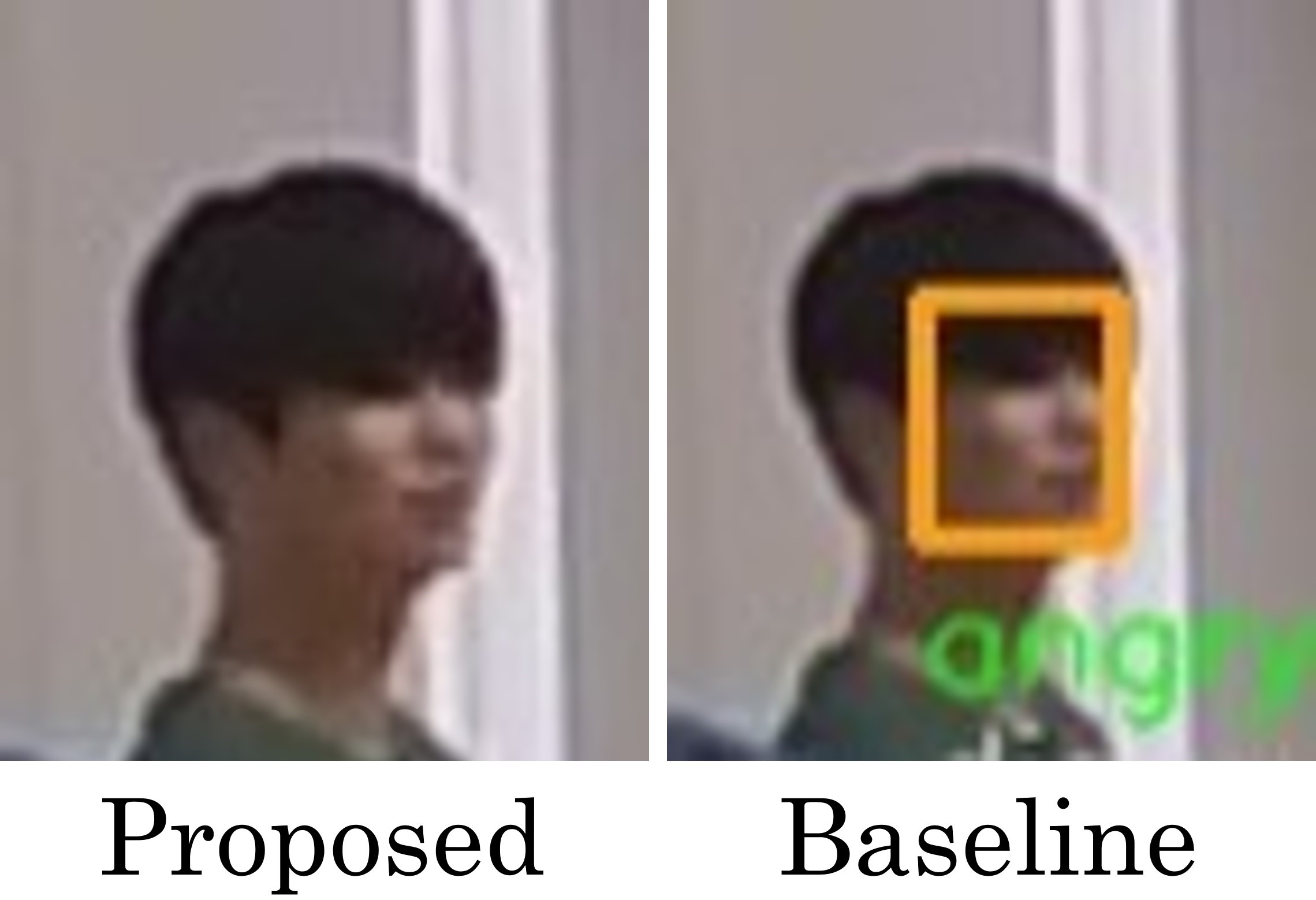}
\subcaption{Difference in detection between (a) and (c)}
\label{Ex1d}
\end{minipage}
\caption{Results in Experiment 1.}
\label{Ex.1}
\end{center}
\end{figure}

%EX1幸せ値に関する検証
%
%本節では提案手法がベースライン手法よりも$H$が高いフレームが抽出できるかを実験1を用いて検証する．図8(a)は提案手法で抽出した$H$が最も高い全方位フレームで図8(b)がベースライン手法で抽出した$H$が最も高い全方位フレームである．図8(d)は図8(a)より生成した透視投影画像で，図8(e)は図8(b)より生成した透視投影画像である．これらの画像を生成するのに用いたパラメータ$(s, t)$と$(w_p, h_p)$を表3に示し，$H$も同様にEx.1は提案手法が0.458, ベースライン手法が0.400と提案手法の方が0.058高いことがわかる．
In Experiment 1, using the scene where six people are the photographed participants and four people are not related, we validate whether the proposed method can extract frames with higher $H$ than the baseline method. Figures \ref{Ex.1}(a) and (b) show the omni-directional frames with the highest $H$ extracted by the proposed method and the baseline method, respectively. Figures \ref{Ex.1}(d) and (e) are the perspective projection images generated from Figs. \ref{Ex.1}(a) and (b), and $H$ of the images by the proposed and baseline methods are 0.458 and 0.400, respectively, indicating that the proposed method is 0.058 higher. 
This is because the proposed method was able to compute $H$ for all the frames by setting the upper and lower sizes of the bounding box and eliminating false detections and interpolating undetected participants. On the other hand, for the baseline method, only 37.2\% of all frames had a mode, as shown in Table \ref{baseline_detection}. For example, as shown in Figs. \ref{Ex.1}(c) and (f), in the frame extracted by the proposed method, an unrelated person in the distance was detected by the baseline method, and this frame was not extracted because the number of detections is seven.
%今後の記述量に合わせて，細かいとこまで書くかどうか決める

\begin{figure}[t!]
\begin{center}
\begin{minipage}{0.32\textwidth}
\includegraphics[width=\textwidth]{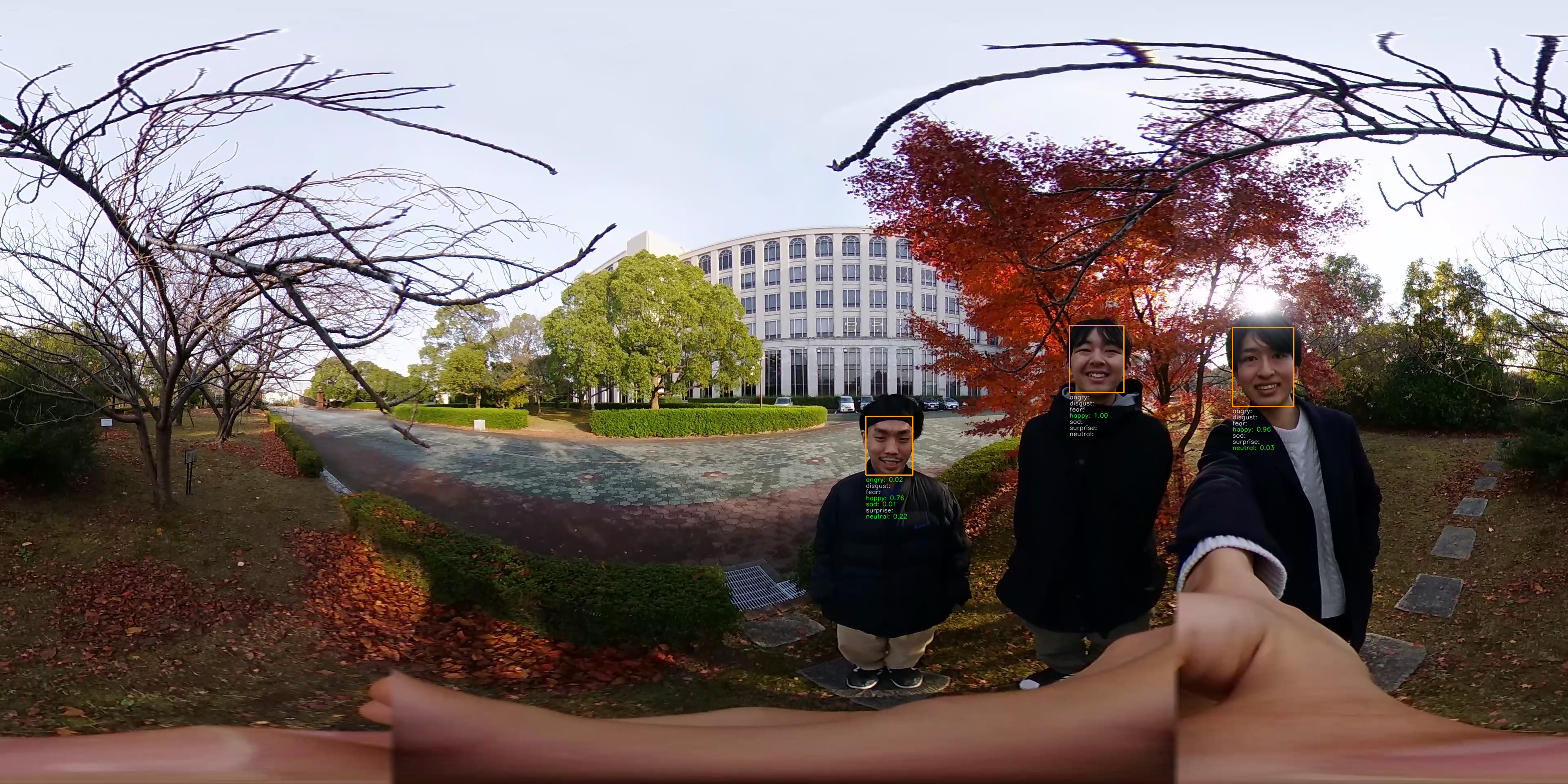}
\subcaption{Frame with highest $H$ by the proposed method} 
\label{Ex2pF}
\end{minipage}
\begin{minipage}{0.32\textwidth}
\includegraphics[width=\textwidth]{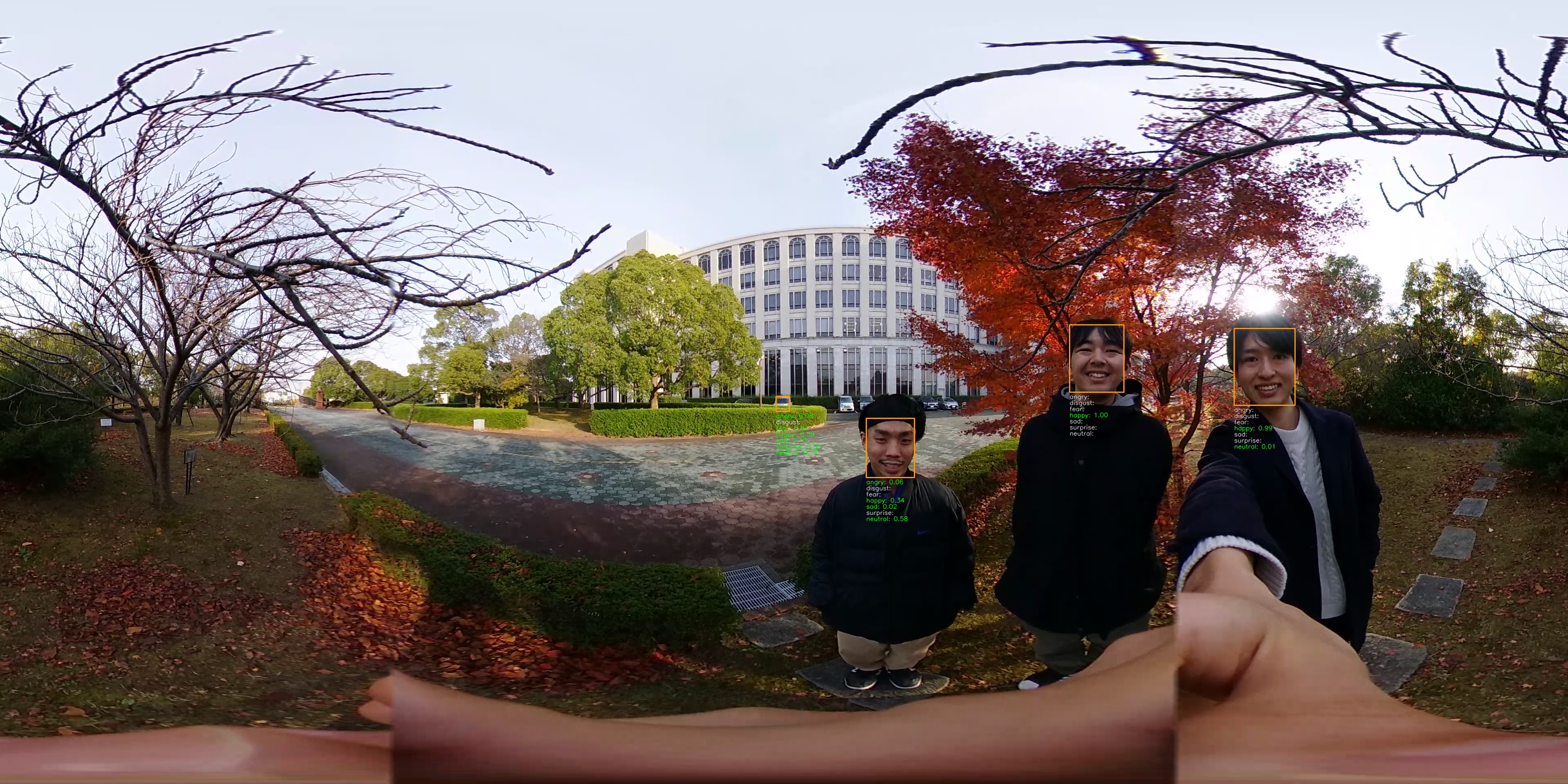}
\subcaption{Frame with highest $H$ by the baseline method} 
\label{Ex2bF}
\end{minipage}
\begin{minipage}{0.26\textwidth}
\includegraphics[width=\textwidth]{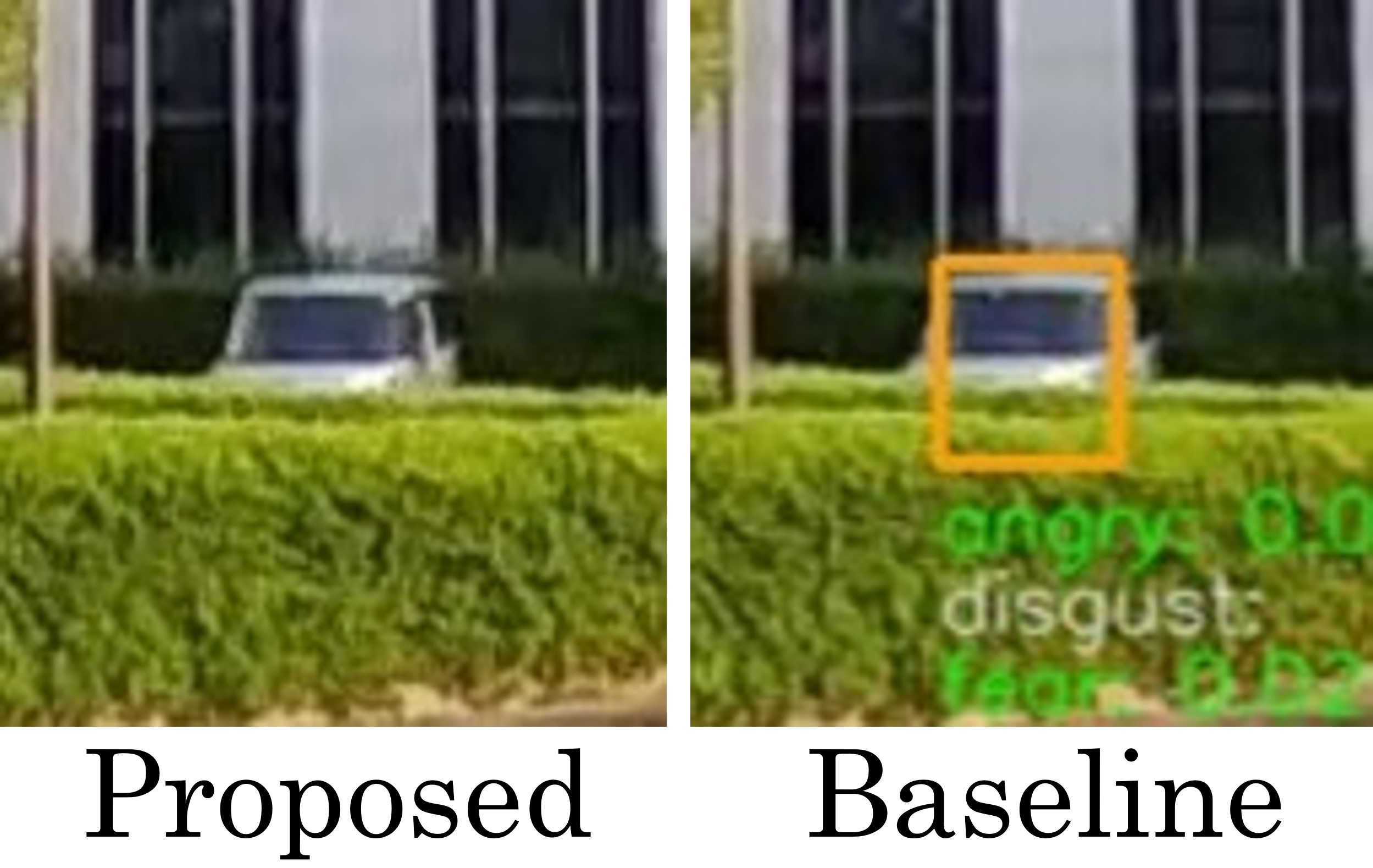}
\subcaption{Comparison of false detection}
\label{Ex2d}
\end{minipage}
\begin{minipage}{0.34\textwidth}
\includegraphics[width=\textwidth]{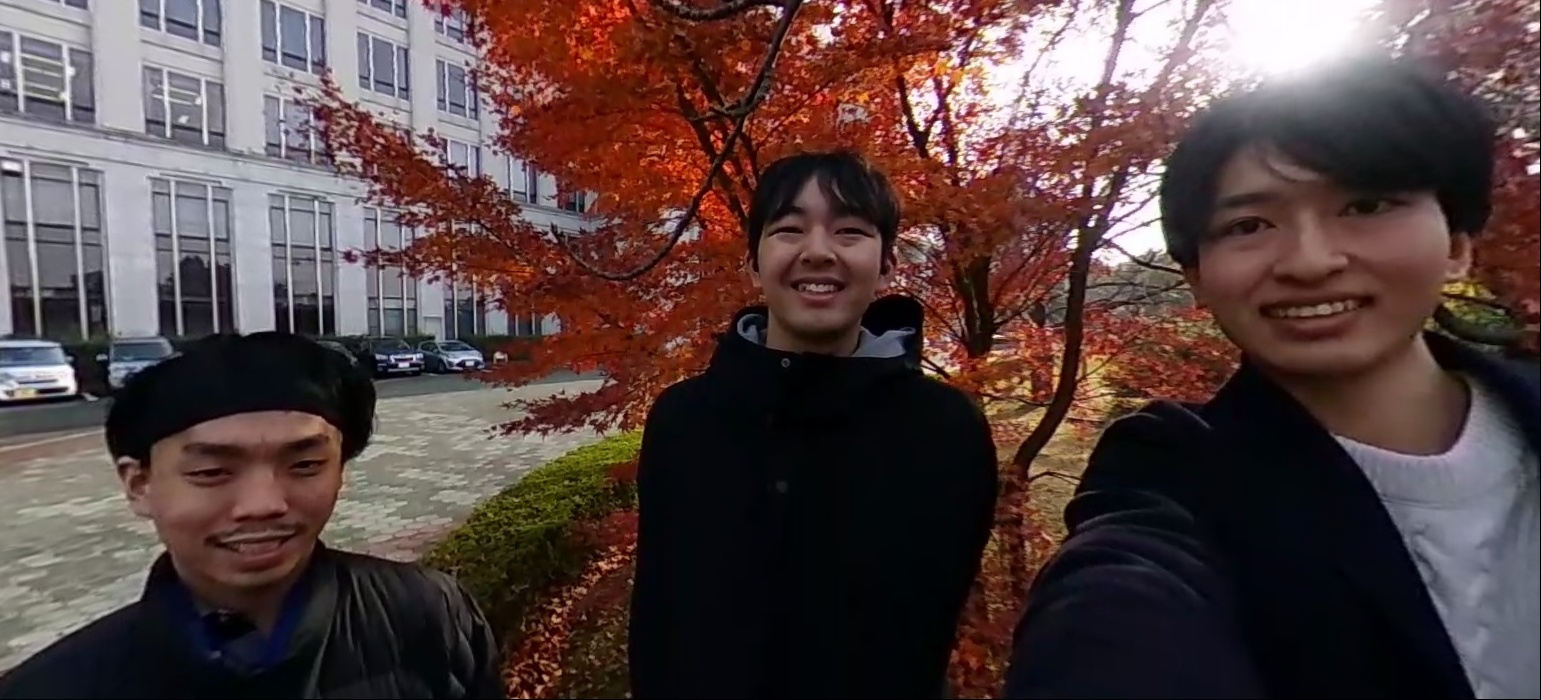}
\subcaption{Generated image from (a)}
\label{Ex2pp}
\end{minipage}
\begin{minipage}{0.35\textwidth}
\includegraphics[width=\textwidth]{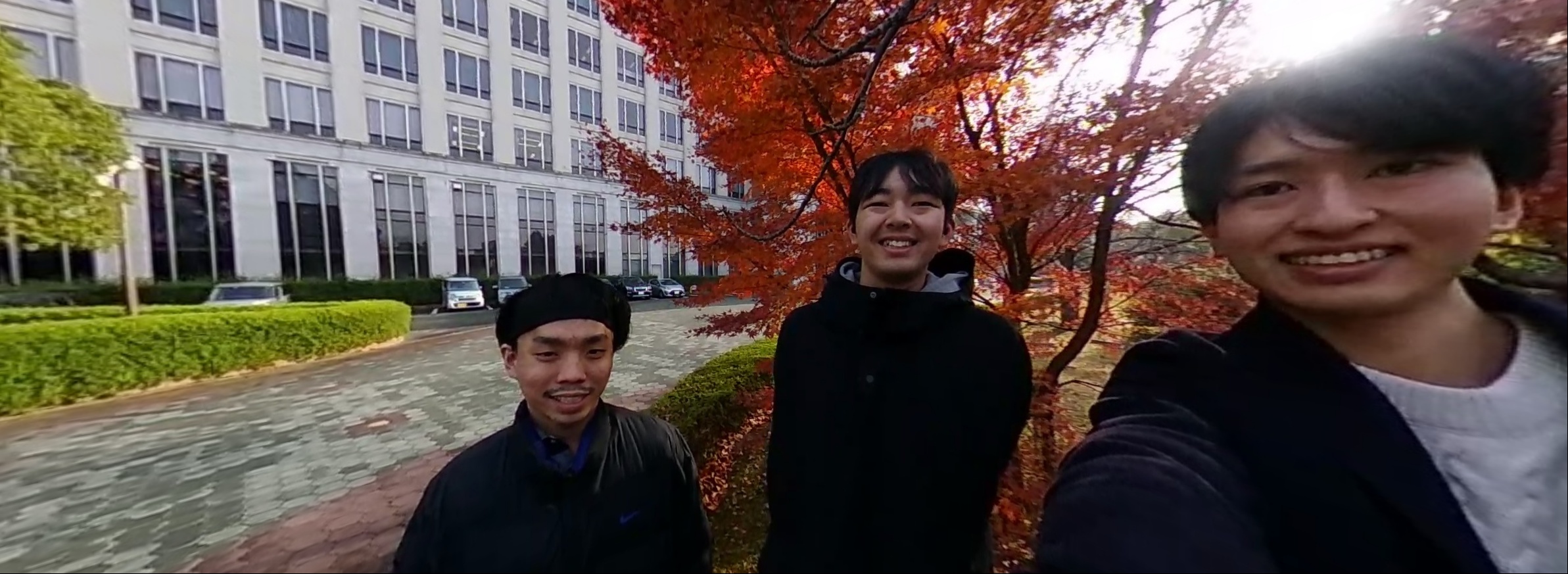}
\subcaption{Generated image from (b)}
\label{Ex2bp}
\end{minipage}
\caption{Results in Experiment 2.}
\label{Ex.2}
\end{center}
%\end{figure}
%\begin{figure}[h]
\begin{center}
\begin{minipage}{0.32\textwidth}
\includegraphics[width=\textwidth]{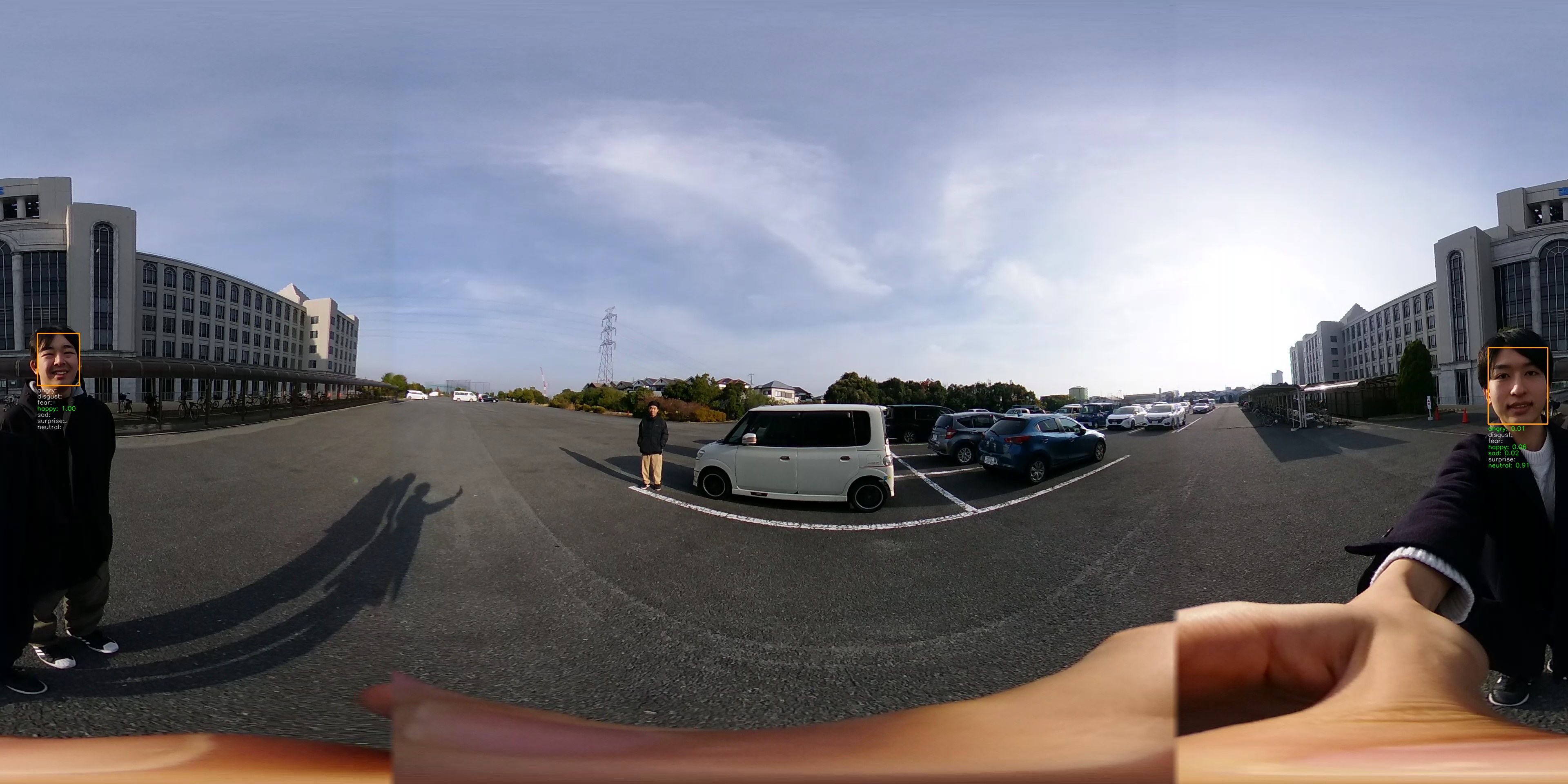}
\subcaption{Frame with highest $H$ by the proposed method} 
\label{Ex3pF}
\end{minipage}
\begin{minipage}{0.32\textwidth}
\includegraphics[width=\textwidth]{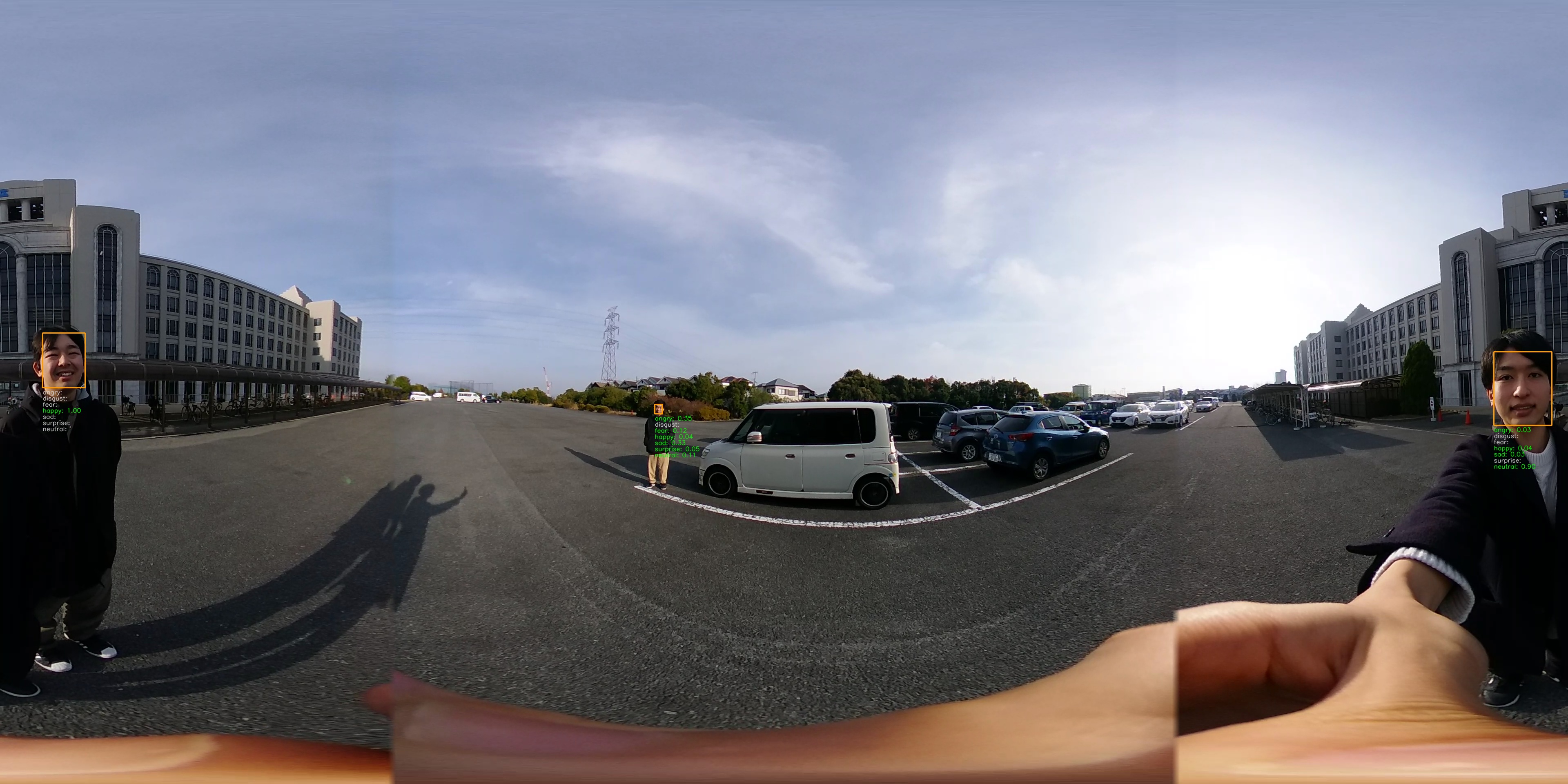}
\subcaption{Frame with highest $H$ by the baseline method} 
\label{Ex3bF}
\end{minipage} 
\begin{minipage}{0.34\textwidth}
\includegraphics[width=\textwidth]{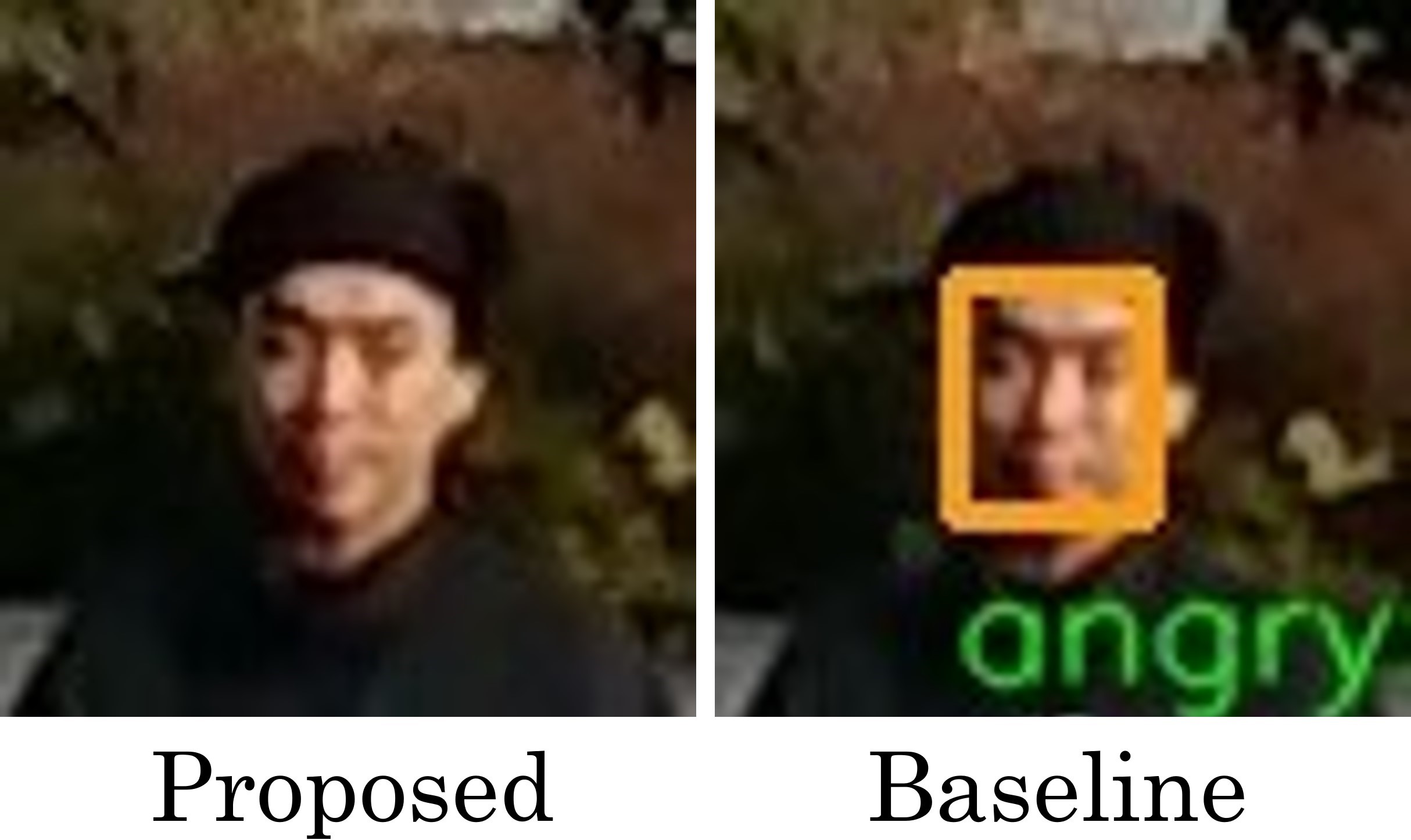}
\subcaption{Comparison of detection} 
\label{Ex3d}
\end{minipage}\\
\begin{minipage}{0.34\textwidth}
\includegraphics[width=\textwidth]{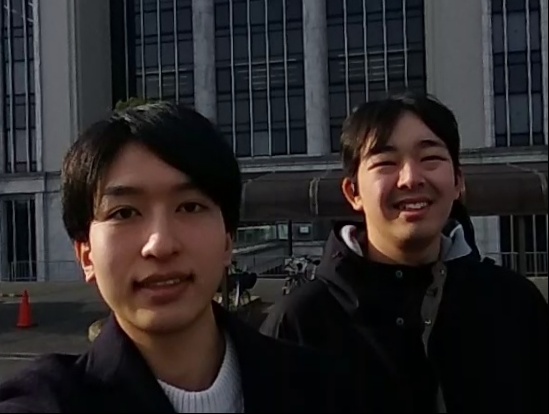}
\subcaption{Generated image from (a)}
\label{Ex3pp}
\end{minipage}
\begin{minipage}{0.64\textwidth}
\includegraphics[width=\textwidth]{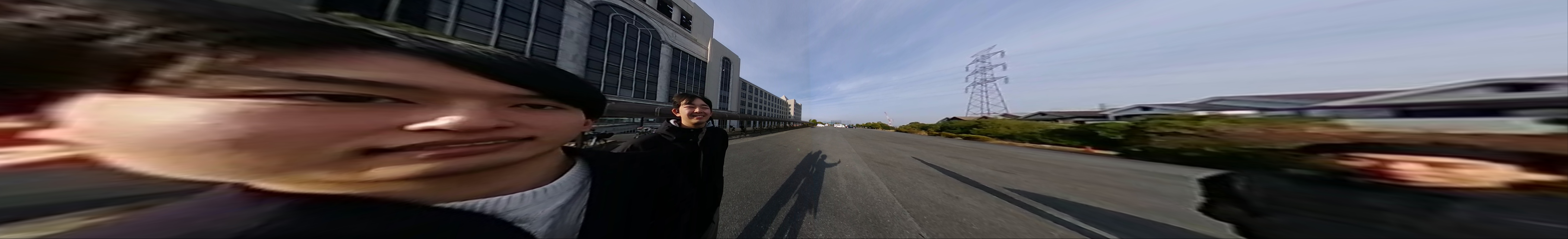}
\subcaption{Generated image from (b)}
\label{Ex3bp}
\end{minipage}
\caption{Results in Experiment 3.}
\label{Ex.3}
\end{center}
\end{figure}

\subsection{Validation on false detection elimination}
In Experiments 2 and 3, we validate whether the elimination of false detections by the proposed method is effective. First, in Experiment 2, although the number of photographed participants is three, the mode by the baseline method is four as shown in Table \ref{baseline_detection}. This resulted in extracting the frame with the highest $H$ (Fig. \ref{Ex.2}(b)), which contains the false detection car shown on the right side of Fig. \ref{Ex.2}(c), and generating the perspective projection image with unnecessarily wide angle of view by the baseline method shown in Fig. \ref{Ex.2}(e). Also, $H$ is calculated as if the car is a face. Such a happiness value is not appropriate for selecting a frame.

%In this section, we verify whether the proposed method of eliminating false detections is more useful than the baseline method by using Experiments 2 and 3. First, we explain Experiment 2 in which the number of photographed participants is 3. The baseline method extracted the frame with the highest $H$, \ref{Ex2bF}, which contains the false detection car shown on the right side of Fig. \ref{Ex2d}. As shown in Table \ref{baseline_detection}, the mode is 4, so the frame with the highest $H$, which is calculated from the happiness value of the false detection car and the happiness values of the three persons is extracted, which is not desirable because $H$ is affected by the false detection. 
%This is not a desirable result because $H$ is affected by false positives. In addition, the generated Fig. \ref{Ex2bp} is determined to be the face with the car farthest to the left, and a perspective projection image that is elongated in the horizontal direction is generated. It was also observed that 101 frames had false detections, 2 frames were undetected, and both occurred at the same time. The most frequent frame was only 52, or 36.3\% of all frames, and the remaining frames were not selected by $H$. 

In contrast, the results of the proposed method shown in Fig. \ref{Ex.2}(a) and the left figure in Fig. \ref{Ex.2}(c) and \ref{Ex.2}(d) show that the false detection of the car is eliminated and the perspective projection image contains only the participants. Specifically, in the procedure of the proposed method, the mean shift classifies the bounding boxes into seven classes, and threshold $T$ for the ratio of detections eliminated four types of false detections, which occurred in 32 frames, as shown in Table \ref{proposed_detection}. As a result, three classes remained, which are number of participants.

Next, we describe Experiment 3, in which the number of people photographed is two and the number of unrelated persons is one. The baseline method extracted the frame with the highest $H$ (Fig. \ref{Ex.3}(b)), in which an unrelated person about four meters away from the camera was detected, as shown on the right side of Fig. \ref{Ex.3}(c). In this case, the angle of view of the perspective projection image that includes the three people becomes considerably larger, and the texture near the edges of the image is significantly stretched as shown in Fig. \ref{Ex.3}(e), which does not have proper visibility.

On the other hand, the results of the proposed method shown in Figs. \ref{Ex.3}(a), \ref{Ex.3}(d) and the left of \ref{Ex.3}(c) show that the false detection was eliminated and that only the target participants appear in the perspective projection images. Since the elimination of the false detection in this case is due to the setting of the minimum size of bounding boxes, we confirmed that such setting is effective. 
From the results, we can say that the proposed method of eliminating false detections by the mean shift clustering is more effective than the baseline method using the mode to generate perspective projection images in which only the photographed participants appear.

\begin{figure}[t!]
\begin{center}
\begin{minipage}{0.36\textwidth}
\includegraphics[width=\textwidth]{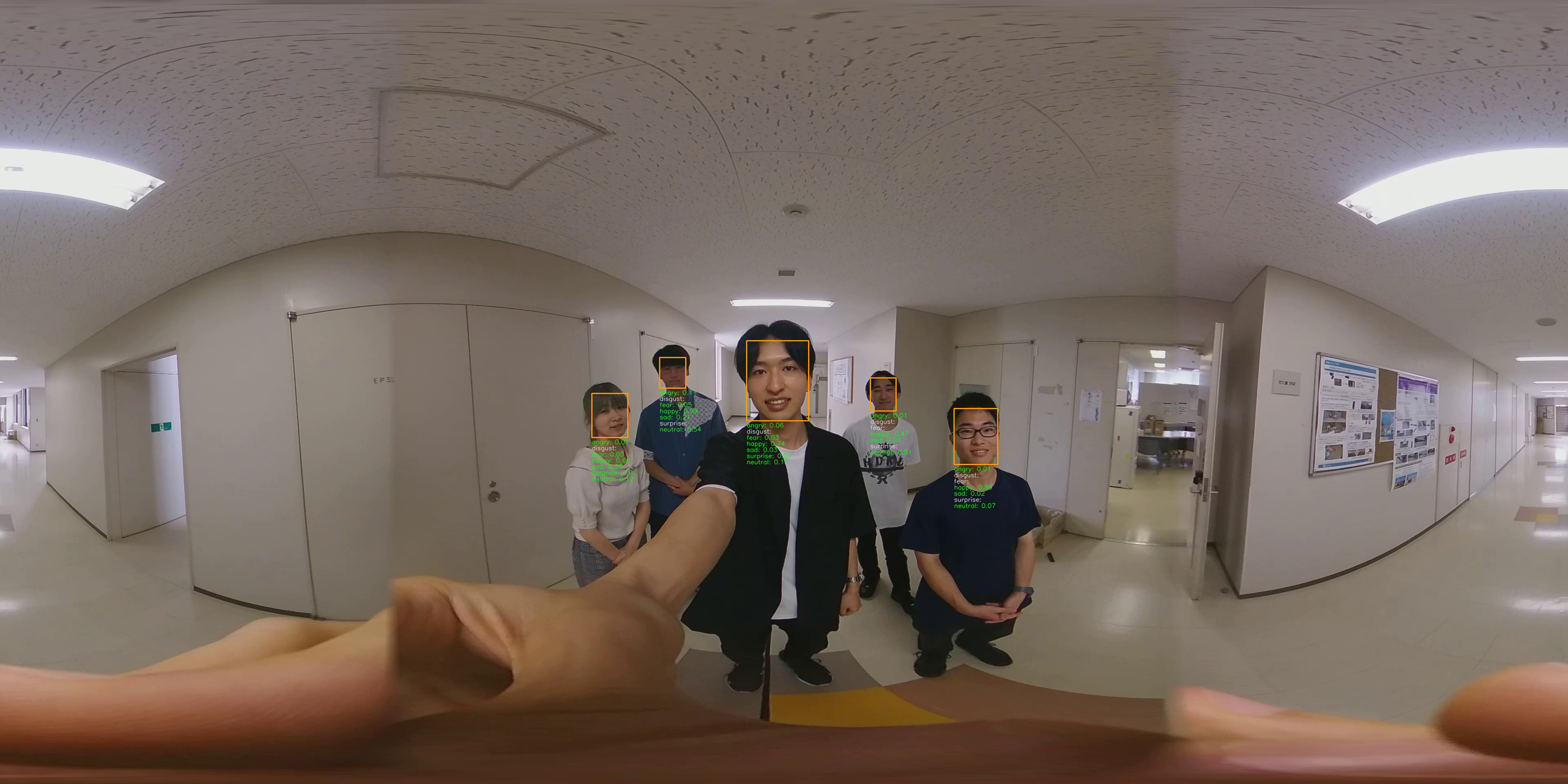}
\subcaption{Frame with highest $H$ by the proposed method} 
\label{Ex4pF}
\end{minipage}
\begin{minipage}{0.36\textwidth}
\includegraphics[width=\textwidth]{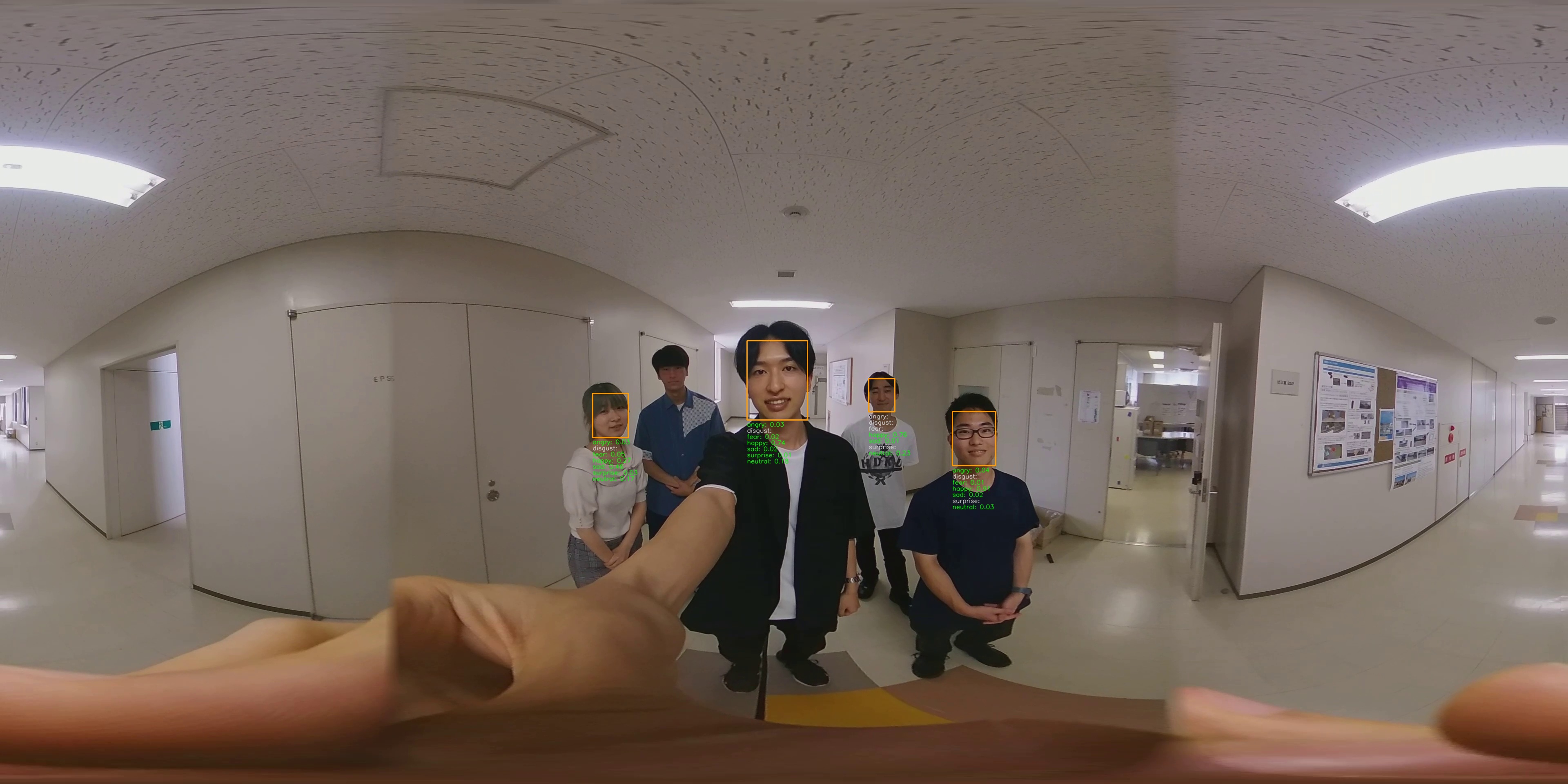}
\subcaption{Frame with highest $H$ by the baseline method} 
\label{Ex4bF}
\end{minipage}
\begin{minipage}{0.22\textwidth}
\includegraphics[width=\textwidth]{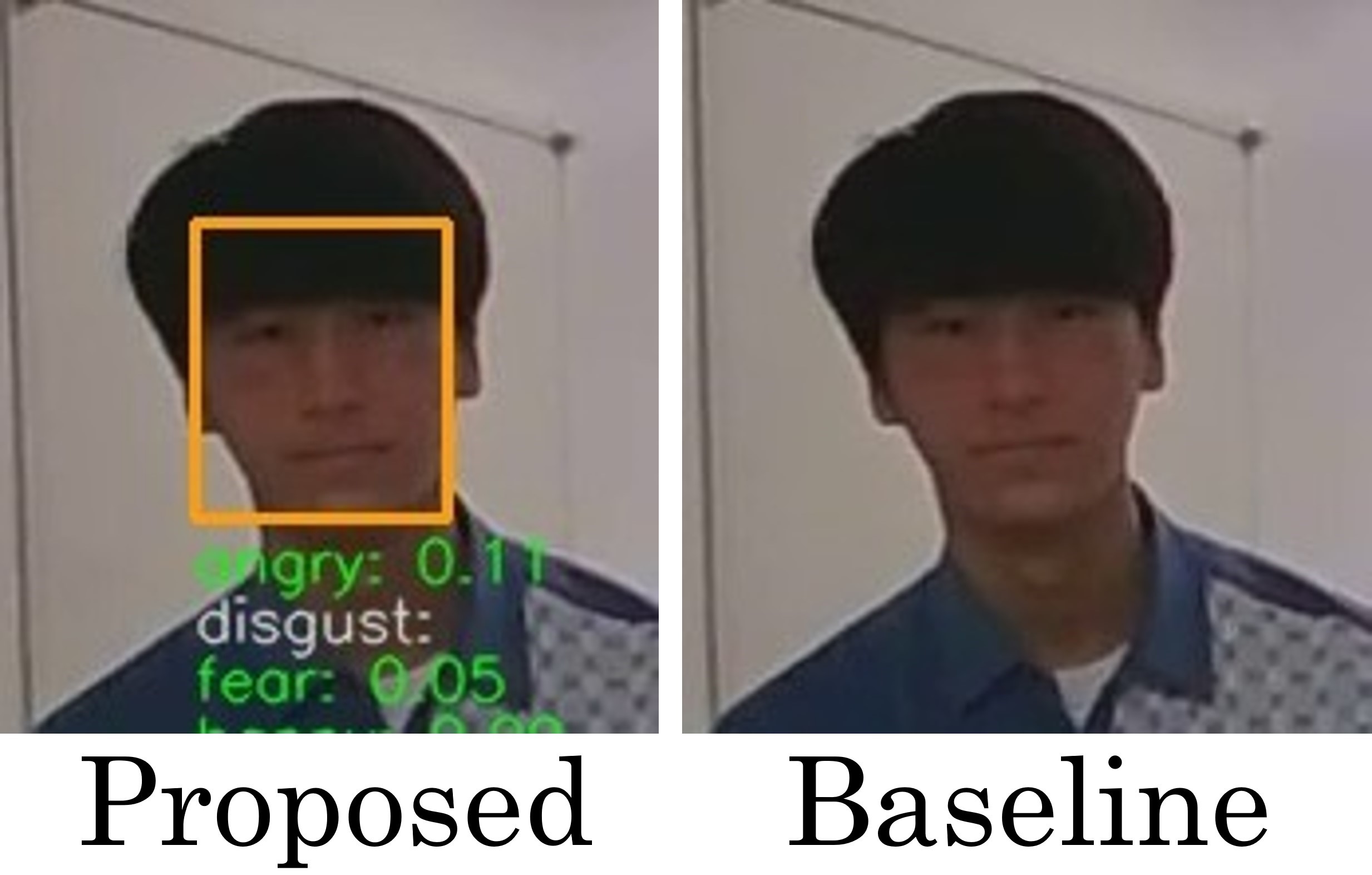}
\subcaption{Comparison of detection}
\label{Ex4d}
\end{minipage}
\begin{minipage}{0.36\textwidth}
\includegraphics[width=\textwidth]{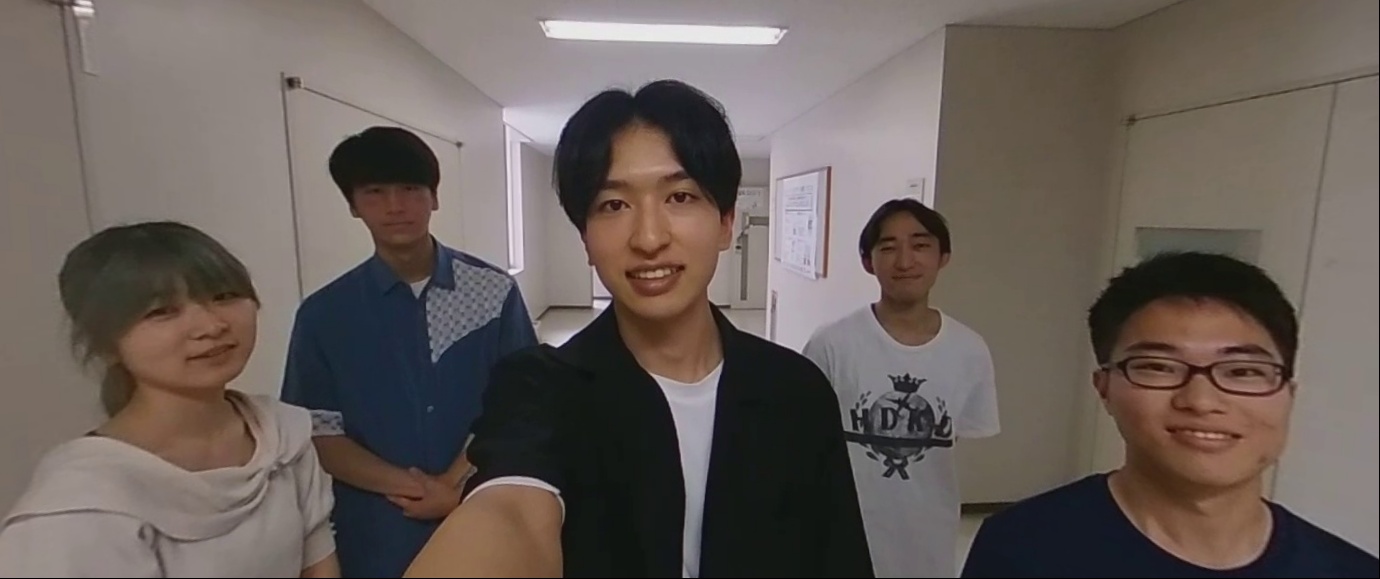}
\subcaption{Generated image from (a)}
\label{Ex4pp}
\end{minipage}
\begin{minipage}{0.36\textwidth}
\includegraphics[width=\textwidth]{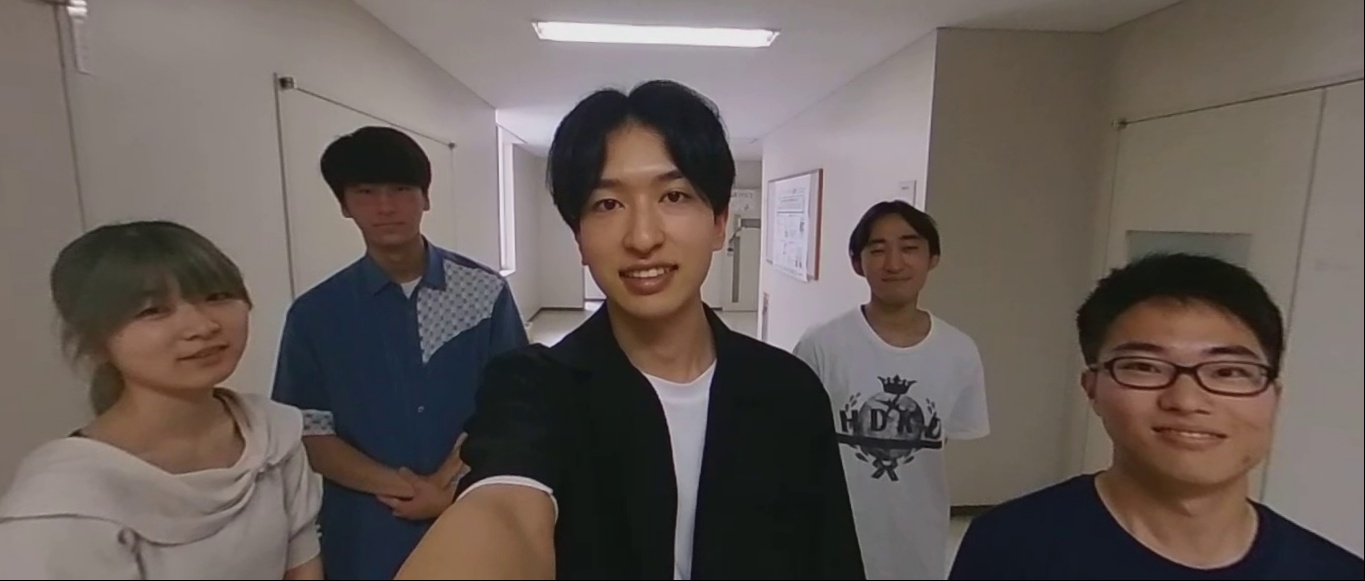}
\subcaption{Generated image from (b)}
\label{Ex4bp}
\end{minipage}
\caption{Results in Experiment 4.}
\label{Ex.4}
\end{center}
%\end{figure}
%\begin{figure}[tb]
\begin{center}
\begin{minipage}{0.36\textwidth}
\includegraphics[width=\textwidth]{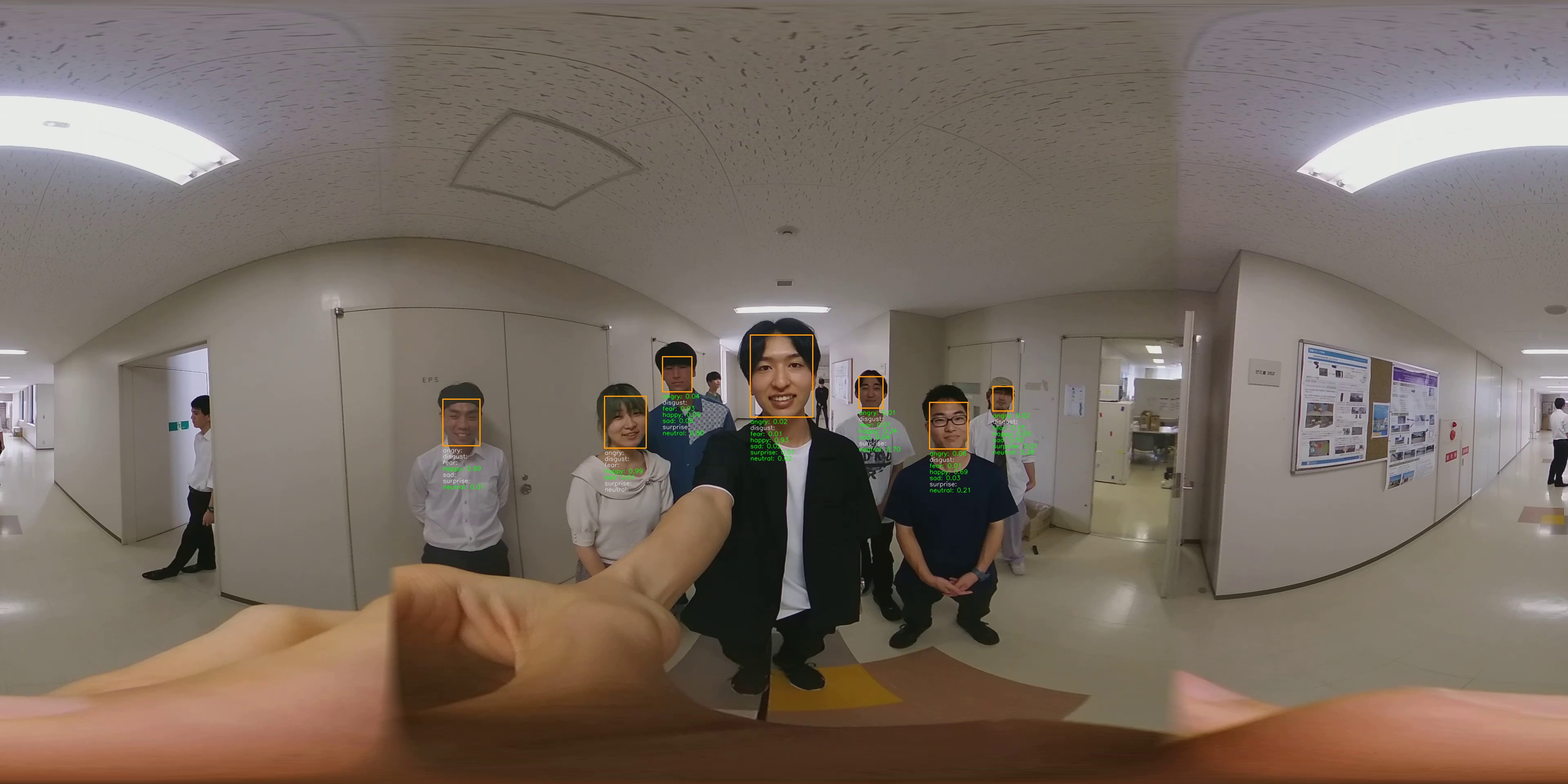}
\subcaption{Frame with highest $H$ by the proposed method} 
\label{Ex5pF}
\end{minipage}
\begin{minipage}{0.36\textwidth}
\includegraphics[width=\textwidth]{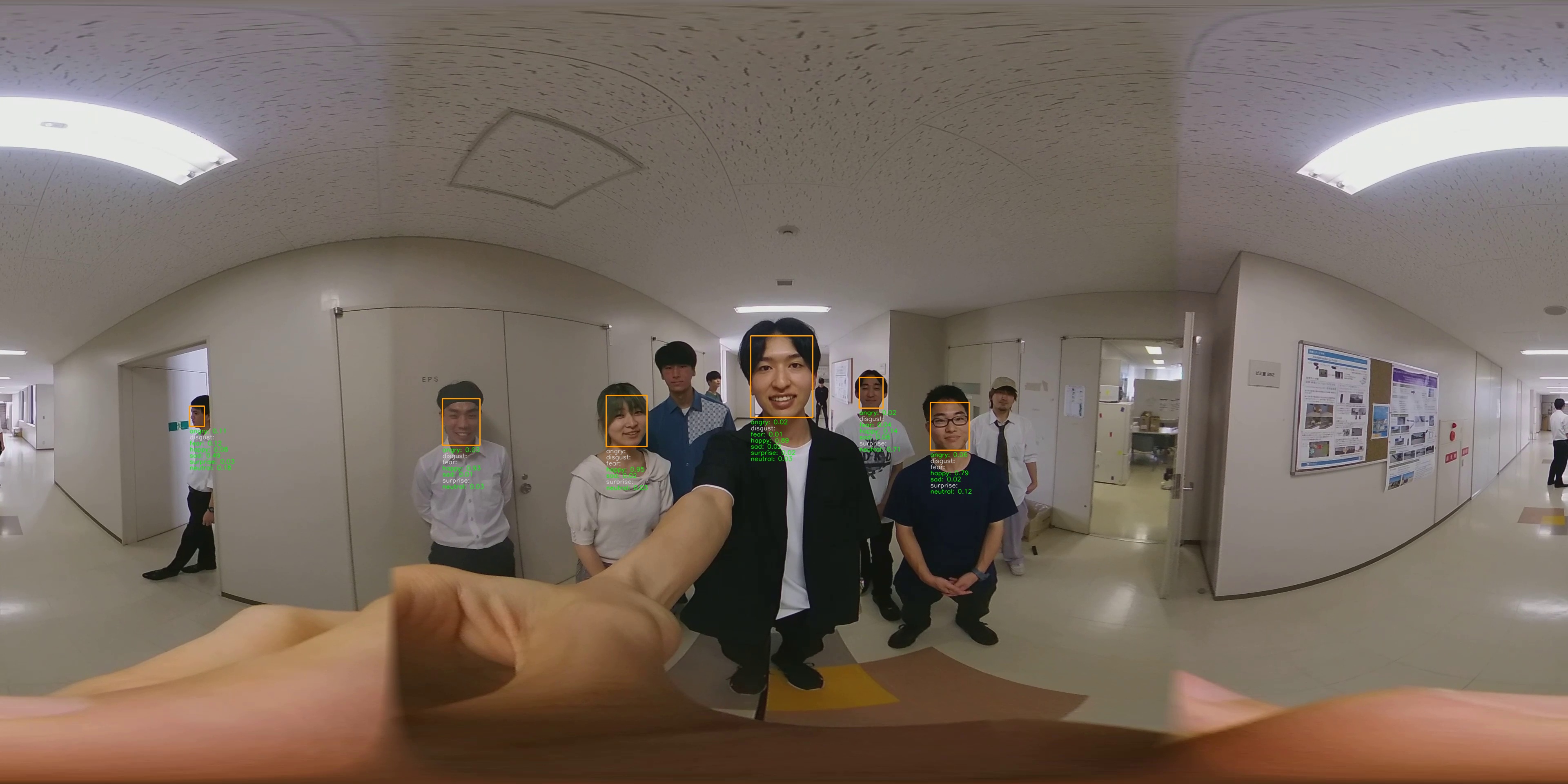}
\subcaption{Frame with highest $H$ by the baseline method} 
\label{Ex5bF}
\end{minipage}
\begin{minipage}{0.49\textwidth}
\includegraphics[width=\textwidth]{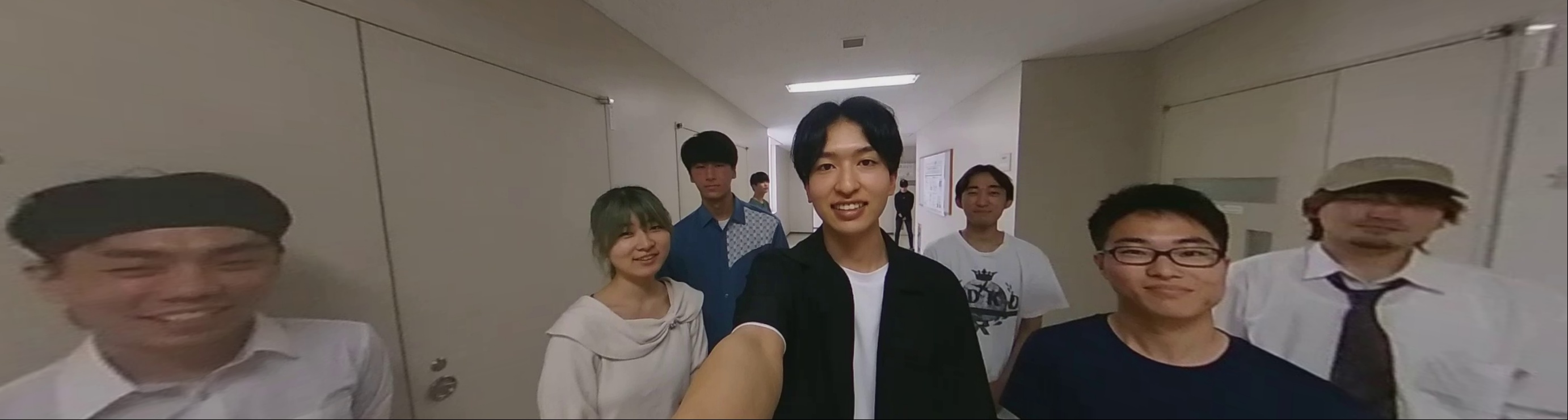}
\subcaption{Generated image from (a)}
\label{Ex5pp}
\end{minipage}
\begin{minipage}{0.49\textwidth}
\includegraphics[width=\textwidth]{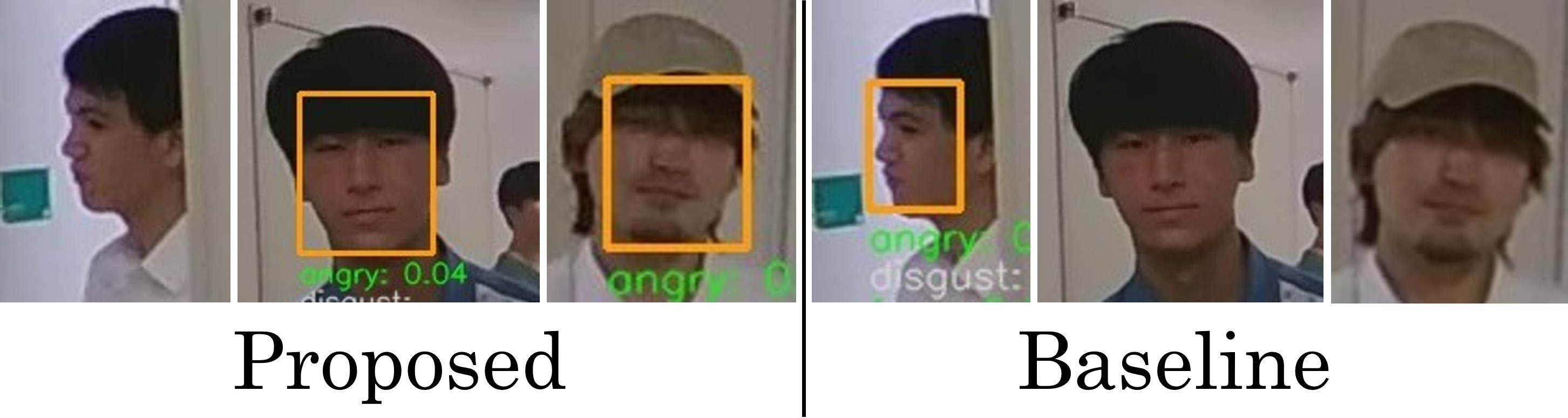}
\subcaption{Comparison of detection} 
\label{Ex5d}
\end{minipage}
\caption{Results in Experiment 5.}
\label{Ex.5}
\end{center}
\end{figure}

\subsection{Validation on interpolation of Undetected Faces}
%本節ではベースライン手法にはない未検出補間を実装した提案手法が有用であるかを実験4・5を用いて検証する．まず，撮影対象人数が5人の実験4から説明する．ベースライン手法ではFig. \ref{Ex4d}の右側に示すような未検出の人物を除いた$H$が最も高いフレームであるFig. \ref{Ex4bF}が抽出された．これはTable \ref{baseline_detection}に示すように最頻値が4となっているため，$H$が未検出を考慮していないため望ましい結果ではない．しかし，生成されたFig. \ref{Ex4bp}は最も左と右の顔の人物が検出されて決定されていることから未検出の人物が写された透視投影画像が出力できているが，未検出の人物の幸せ値を考慮していないため，$H$が最も高いフレームであるとは言えない．また，誤検出が1フレームと少ないが，39フレームで未検出であり，全41フレーム中のほとんどで未検出が発生していることが確認できる．また，最頻値フレームが25フレームと全体の60.9\%であり，残りのフレームが抽出対象外であるのと，撮影対象全員が検出されたフレームが2フレームしかなく，全体の4.8\%であることより，ベースライン手法では全員の幸せ値が最も高いフレームを抽出することができないことが言える．
%対してFigs. \ref{Ex4pF}と\ref{Ex4d}の左側と\ref{Ex4pp}に示す提案手法の結果では，ベースライン手法で未検出だった人物が検出されており，撮影対象者らのみが写された透視投影画像が出力されていることが確認できる．Table \ref{proposed_method}に示すように提案手法でも32フレームの未検出が確認できるが，撮影対象者らそれぞれが全てのフレームで25\%以上検出されていることから，クラスタリング時に取得したバウンディングボックスの平均値を用いて未検出を補間できている．
In Experiments 4 and 5, we validate whether the interpolation of undetected faces is effective. First, let us start with Experiment 4, in which the number of participants is five. The baseline method extracted the frame with the highest $H$ (Fig. \ref{Ex.4}(b)), and output Fig. \ref{Ex.4}(e), excluding undetected persons, as shown on the right side of Fig. \ref{Ex.4}(c). In this scene, the detection accuracy of the some participants by MTCNN was poor, and the mode throughout all the frame is four, as shown in Table \ref{baseline_detection}. Also, all the five participants were detected in only two frames, indicating that the baseline method cannot appropriately extract the frame with the highest $H$ for all the participants.

In contrast, the results of the proposed method shown in Figs. \ref{Ex.4}(a) and (d) and the left of Fig. \ref{Ex.4}(c) show that the proposed method interpolated undetected faces. As shown in Table \ref{proposed_detection}, although the proposed method still failed to detect faces in 32 frames, since each of the five participants was detected in more than 25\% of all the frames, the mean value of the bounding boxes for each person could be used to interpolate the undetected persons.

Next, we describe Experiment 5, in which the number of people photographed is seven and the number of unrelated persons is four. The baseline method extracted the frame with the highest $H$ (Fig. \ref{Ex.5}(b)), which has one false detection and two undetected, as shown in the three images on the right side of Fig. \ref{Ex.5}(d). In this case, the mode is six as shown in Table \ref{baseline_detection}, which indicates that the persons detected in each frame differs depending on false detections and undetections. In addition, since the falsely detected person was located further away from the target participants, the calculated $(s, t)$ i.e., the angle of view, became larger as shown in Table \ref{ExResults}, resulting in failing to generate an appropriately visible image.

%Next, we describe Experiment 5, in which the number of people photographed is seven and the number of unrelated persons is four. The baseline method extracted the frame with the highest $H$, Fig. \ref{Ex5bF}, which has one false detection and two undetected, as shown in the three frames on the right side of Fig. \ref{Ex5d}. This frame has a mode of 6, as shown in Table \ref{baseline_detection}, which also indicates that the person detected in each frame differs depending on false detections and undetections. In addition, as shown in Table \ref{ExResults}, $(s, T)$ used to calculate the angle of view is very large at $(145.4, 34.72)$, and the angle of view is close to 180°, resulting in failure to generate a perspective projection image. The false detections were 34 frames and the undetected were 32 frames, which accounted for a large percentage of the 35 frames, and none of the frames had neither false detections nor undetected. This indicates that the baseline method is inadequate to deal with false detections and undetections, and is unable to extract the frames with the highest happy values for everyone. 
In contrast, the results of the proposed method shown in Figs. \ref{Ex.5}(a), (c) and the left three images in Fig. \ref{Ex.5}(d) show that the false detection was eliminated and the undetected persons were interpolated. As a result, only the participants appear in the perspective projection images. As shown in Table \ref{proposed_detection}, even though some of participants were undetected in 30 frames in the proposed method, the mean-shift clustering determined the number of classes appropriate for the number of participants, and all undetected faces were interpolated. From these results, we confirmed that the proposed method with undetected interpolation is effective because it can calculate $H$ for all the photographed participants in all the frames.

\section{Conclusion}
%本研究では，視覚障害者が自撮りする場面に着目し，全方位カメラを用いて複数人で自撮りする手法を提案した．提案手法では，全フレーム間の一貫性を考慮して誤検出の排除と未検出の顔の補間をして表情認識を行った後，最終的に撮影対象者らの表情が最も幸せなフレームを抽出し，全員がいる透視投影画像の生成に成功した．
%実験では，提案手法の有用性を示すためにベースライン手法との比較を行った．最頻値を用いたベースライン手法では抽出対象のフレーム数が限られたが，誤検出を排除し未検出を補間した提案手法では全てのフレームから最も表情が幸せなフレームを抽出することに成功した．
%今後の課題として，誤検出や未検出を減らすために別の顔検出モデルを用いるなど，経験的に決めたパラメータを用いている誤検出排除と未検出補間を強化することが挙げられる．
In this study, we proposed a method that enables to take selfies with multiple people using an omni-directional camera. The proposed method succeeded in eliminating false detections and interpolating undetected faces to extract the frame with the happiest expressions of all the photographed participants from all the frames, and generating a perspective projection image in which all the participants are present.
In the experiments, we compared the proposed method with the baseline method to demonstrate its effectiveness. 
Future work includes the determination of optical parameters for enhancing false detection elimination and undetected interpolation.

%articles~\cite{ref_article1}, an LNCS chapter~\cite{ref_lncs1}, a
%book~\cite{ref_book1}, proceedings without editors~\cite{ref_proc1},
%and a homepage~\cite{ref_url1}. Multiple citations are grouped
%\cite{ref_article1,ref_lncs1,ref_book1},
%\cite{ref_article1,ref_book1,ref_proc1,ref_url1}.
\bibliographystyle{splncs04}
\bibliography{egbib}

\begin{thebibliography}{10}
\providecommand{\url}[1]{\texttt{#1}}
\providecommand{\urlprefix}{URL }
\providecommand{\doi}[1]{https://doi.org/#1}

\bibitem{BeMyEyes}
Be my eyes. \url{https://www.bemyeyes.com/}, (Accessed on 24/11/2023)

\bibitem{EnvisionAI}
Envision ai. \url{https://apps.apple.com/jp/app/envision-ai/id1268632314}, (Accessed on 24/11/2023)

\bibitem{OrCamMyEye2}
Orcam myeye 2. \url{https://www.orcam.com/en/myeye2/}, (Accessed on 24/11/2023)

\bibitem{TapTapSee}
Taptapsee. \url{https://taptapseeapp.com/}, (Accessed on 24/11/2023)

\bibitem{balata2015blindcamera}
Balata, J., Mikovec, Z., Neoproud, L.: Blindcamera: Central and golden-ratio composition for blind photographers. In: Proceedings of the Mulitimedia, Interaction, Design and Innnovation, pp.~1--8 (2015)

\bibitem{bigham2010vizwiz}
Bigham, J.P., Jayant, C., Ji, H., Little, G., Miller, A., Miller, R.C., Miller, R., Tatarowicz, A., White, B., White, S., et~al.: Vizwiz: nearly real-time answers to visual questions. In: Proceedings of the 23nd annual ACM Symposium on User Interface Software and Technology. pp. 333--342 (2010)

\bibitem{comaniciu2002mean}
Comaniciu, D., Meer, P.: Mean shift: A robust approach toward feature space analysis. IEEE Transactions on Pattern Analysis and Machine Intelligence  \textbf{24}(5),  603--619 (2002)

\bibitem{cutter2015towards}
Cutter, M.P., Manduchi, R.: Towards mobile ocr: How to take a good picture of a document without sight. In: Proceedings of the 2015 ACM Symposium on Document Engineering. pp. 75--84 (2015)

\bibitem{gonzalez2022understanding}
Gonzalez~Penuela, R.E., Vermette, P., Yan, Z., Zhang, C., Vertanen, K., Azenkot, S.: Understanding how people with visual impairments take selfies: Experiences and challenges. In: Proceedings of the 24th International ACM SIGACCESS Conference on Computers and Accessibility. pp.~1--4 (2022)

\bibitem{iwamura2020visphoto}
Iwamura, M., Hirabayashi, N., Cheng, Z., Minatani, K., Kise, K.: Visphoto: photography for people with visual impairment as post-production of omni-directional camera image. In: Extended Abstracts of the 2020 CHI Conference on Human Factors in Computing Systems. pp.~1--9 (2020)

\bibitem{jayant2011supporting}
Jayant, C., Ji, H., White, S., Bigham, J.P.: Supporting blind photography. In: Proceedings of the 13th International ACM SIGACCESS Conference on Computers and Accessibility. pp. 203--210 (2011)

\bibitem{jokela2019people}
Jokela, T., Ojala, J., V{\"a}{\"a}n{\"a}nen, K.: How people use 360-degree cameras. In: Proceedings of the 18th International Conference on Mobile and Ubiquitous Multimedia. pp. 1--10 (2019)

\bibitem{kacorri2017people}
Kacorri, H., Kitani, K.M., Bigham, J.P., Asakawa, C.: People with visual impairment training personal object recognizers: Feasibility and challenges. In: Proceedings of the 2017 CHI Conference on Human Factors in Computing Systems. pp. 5839--5849 (2017)

\bibitem{kutiyanawala2011teleassistance}
Kutiyanawala, A., Kulyukin, V., Nicholson, J.: Teleassistance in accessible shopping for the blind. In: Proceedings on the International Conference on Internet Computing. p.~1 (2011)

\bibitem{vOICe}
Meijer, P.: The voice - new frontiers in sensory substitution. https://www.seeingwithsound.com/, (Accessed on 24/11/2023)

\bibitem{obata2020asynchronous}
Obata, K., Nakamura, Y., Chen, L., Augeri, J.: Asynchronous co-eating through video message exchange: Support for making video messages. In: Proceedings of Cross-Cultural Design. Applications in Health, Learning, Communication, and Creativity. pp. 338--348 (2020)

\bibitem{justin_shenk_2021_5362356}
Shenk, J., CG, A., Arriaga, O., Owlwasrowk: justinshenk/fer: Zenodo (Sep 2021), \url{https://doi.org/10.5281/zenodo.5362356}, (Accessed on 18/7/2023)

\bibitem{vazquez2014assisted}
V{\'a}zquez, M., Steinfeld, A.: An assisted photography framework to help visually impaired users properly aim a camera. ACM Transactions on Computer-Human Interaction  \textbf{21}(5),  1--29 (2014)

\bibitem{white2010easysnap}
White, S., Ji, H., Bigham, J.P.: Easysnap: real-time audio feedback for blind photography. In: Adjunct Proceedings of the 23nd annual ACM Symposium on User Interface Software and Technology. pp. 409--410 (2010)

\bibitem{zhang2016joint}
Zhang, K., Zhang, Z., Li, Z., Qiao, Y.: Joint face detection and alignment using multitask cascaded convolutional networks. IEEE Signal Processing Letters  \textbf{23}(10),  1499--1503 (2016)

\end{thebibliography}
%Author, F.: Article title. Journal \textbf{2}(5), 99--110 (2016)
%Author, F., Author, S.: Title of a proceedings paper. In: Editor,
%F., Editor, S. (eds.) CONFERENCE 2016, LNCS, vol. 9999, pp. 1--13.
%Springer, Heidelberg (2016). \doi{10.10007/1234567890}
%Author, F., Author, S., Author, T.: Book title. 2nd edn. Publisher,
%Author, A.-B.: Contribution title. In: 9th International Proceedings
%on Proceedings, pp. 1--2. Publisher, Location (2010)
%LNCS Homepage, \url{http://www.springer.com/lncs}. Last accessed 4
%\end{thebibliography}
\end{document}